\documentclass[11pt]{article}

\usepackage[preprint]{acl}

\usepackage{times}
\usepackage{latexsym}

\usepackage[T1]{fontenc}

\usepackage[utf8]{inputenc}

\usepackage{microtype}

\usepackage{inconsolata}

\usepackage{graphicx}

\usepackage{hyperref}       
\usepackage{url}
\usepackage{booktabs}       
\usepackage{amsfonts}       
\usepackage{nicefrac}       
\usepackage{xcolor}
\usepackage{fontawesome5}

\usepackage{amsmath,amsfonts,bm}

\def\eqref#1{equation~\ref{#1}}

\def\1{\bm{1}}

\DeclareMathAlphabet{\mathsfit}{\encodingdefault}{\sfdefault}{m}{sl}
\SetMathAlphabet{\mathsfit}{bold}{\encodingdefault}{\sfdefault}{bx}{n}

\usepackage{pifont}
\newcommand{\xmark}{\ding{55}}

\usepackage{longtable}
\usepackage{colortbl}
\usepackage{soul}
\usepackage{multirow}
\usepackage[textsize=scriptsize]{todonotes}
\usepackage{wrapfig}
\usepackage{caption}  
\usepackage{varwidth}

\newcommand{\hlblue}[1]{\sethlcolor{cyan}\hl{#1}}
\newcommand{\hlgreen}[1]{\sethlcolor{lime}\hl{#1}}
\newcommand{\hlpink}[1]{\sethlcolor{pink}\hl{#1}}
\newcommand{\hlyellow}[1]{\sethlcolor{yellow}\hl{#1}}   

\newcommand{\claimant}{patient}
\newcommand{\RHExample}[6]{
    \textbf{#1} \vspace{0.3em}
    \newline
    \textbf{Post (#6):} 
    #2 
    \vspace{0.3em}
    \newline
    \textbf{Population:} \fcolorbox{black}{orange!20}{#3}
    \newline
    \textbf{Intervention:} \fcolorbox{black}{magenta!20}{#4}
    \newline
    \textbf{Outcome:} \fcolorbox{black}{cyan!20}{#5}
    \vspace{0.3em}
}

\newcommand{\ClaimExample}[3]{
    \textbf{#1} \vspace{0.3em}
    \newline
    \textbf{Post (#3):} 
    #2 
}

\title{Decide less, communicate more: On the construct validity of end-to-end fact-checking in medicine}

\author{Sebastian Joseph$^{1}$\footnotemark[1]\ \ \ \
Lily Chen$^{2}$\thanks{Equal Contribution}\ \ \ \ 
Barry Wei$^{3}$\ \ \ \ 
Michael Mackert$^{1}$\\
\textbf{Iain J.\ Marshall}$^{4}$\ \ \ \
\textbf{Paul Pu Liang}$^{5}$\ \ \ \ 
\textbf{Ramez Kouzy}$^{6}$\ \ \ \ 
\textbf{Byron C.\ Wallace}$^{7}$\ \ \ \ 
\textbf{Junyi Jessy Li}$^{1}$
\\
$^1$The University of Texas at Austin,
$^2$Stanford University,
$^3$Indiana University School of Medicine\\
$^4$King's College London,
$^5$Massachusetts Institute of Technology\\
$^6$The University of Texas MD Anderson Cancer Center, $^7$Northeastern University\\
{\small \tt \{sebaj, mackert, jessy\}@utexas.edu, l1ly@stanford.edu, barrwei@iu.edu, iain.marshall@kcl.ac.uk} \\
{\small \tt ppliang@mit.edu, rkouzy@mdanderson.org, b.wallace@northeastern.edu}
}

\begin{document}
\maketitle

\begin{abstract}

Technological progress has led to concrete advancements in tasks that were regarded as challenging, such as automatic fact-checking. Interest in adopting these systems for public health and medicine has grown due to the high-stakes nature of medical decisions and challenges in critically appraising a vast and diverse medical literature.  Evidence-based medicine connects to every individual, and yet the nature of it is highly technical, rendering the medical literacy of majority users inadequate to sufficiently navigate the domain. Such problems with medical communication ripen the ground for end-to-end fact-checking agents: check a claim against current medical literature and return with an evidence-backed verdict. And yet, such systems remain largely unused.

In this position paper, developed with expert input,
we present the first study examining how clinical experts verify real claims from social media by synthesizing medical evidence. 
In searching for this upper-bound, we reveal fundamental challenges in end-to-end fact-checking when applied to medicine: Difficulties connecting claims in the wild to scientific evidence in the form of clinical trials; ambiguities in underspecified claims mixed with mismatched intentions; and inherently subjective veracity labels. 
We argue that fact-checking should be approached as an interactive communication problem, rather than an end-to-end process. Data and code are available at \url{https://github.com/SebaJoe/decide-less-communicate-more}. 
\end{abstract}

\section{Introduction}

Decision making in medicine is personal, intimate, and high-stakes. 
Traditionally the patient---often a layperson unfamiliar with medicine---converses with their care providers about questions about their health. 
However, the reality is far from this picture: most Americans resort to the web when they have a health-related question \citep{fox2013health}. Today, social media and AI have made medical knowledge seemingly accessible. 
But claims made by others on the web (or by a chatbot) can be inaccurate or inapplicable. 
This, combined with eroding health literacy \citep{champlin2017toward}, has led to challenges in public health~\citep{Hassan2015TheQT} as well as patient-provider sessions. 

Meanwhile, evidence-based medicine has continuously evolved,  with an evidence base growing too rapidly for physicians to keep up \citep{bastian2010seventy,marshall2021state}. 
This provides a unique opportunity for AI fact-checking systems:
Advances in retrieval systems and Large Language Models (LLMs) have increased interest in fact-checking systems that can classify medical claims as `True' or `False' with supporting evidence. 
However, despite technological advances, such systems remain underutilized as they struggle to address diverse \textit{claims in the wild}---naturally occurring statements, usually made by laypeople, that pervade public discourse~\citep{DAS2023103219, chen-etal-2022-generating}; this further applies to claims made by AI agents that can be sometimes questionable. 
Existing fact-checking datasets often miss such claims because they were collected from already curated sources such as fact-checking websites and news articles~\citep{kotonya2020explainableautomatedfactcheckingpublic, vladika-etal-2024-healthfc}. Prior datasets also extracted claims from their context~\citep{sarrouti-etal-2021-evidence-based}, generated synthetic claims~\citep{saakyan-etal-2021-covid}, or filtered claims based on lexical criteria~\citep{mohr-etal-2022-covert}. As a result, systems trained on these heavily curated datasets are likely to fail to understand real medical claims made by the public.

\begin{table*}[t]
\tiny 
    \centering
    \begin{tabular}{>
{\raggedright\arraybackslash}p{1.5cm} |>{\raggedright\arraybackslash}p{1.41cm} >
{\raggedright\arraybackslash}p{1.42cm} >
{\raggedright\arraybackslash}p{2.2cm} >{\raggedright\arraybackslash}p{1.2cm}>{\raggedright\arraybackslash}p{1.3cm} >
{\raggedright\arraybackslash}p{2cm}  }
    
    \toprule
    \textbf{Dataset} & \textbf{Domains} & \textbf{Source} & \textbf{Labels} & \textbf{Explanations} & \textbf{Evidence Type} & \textbf{Claim Example}\\  \midrule
             \sc pubhealth (2020) & Public Health & Fact-Checking Websites, health news & True, Unproven, False, Mixture & \xmark & Sentences from same claim article source & Expired boxes of cake and pancake mix
are dangerously toxic.\\
\midrule
        \sc scifact (2020) & Science  & Expert-written & Supports, Refutes & \xmark & Scientific Articles & Rapamycin slows aging in fruit flies.\\
        \midrule
        \sc healthver (2021) & COVID-19 & News Articles, blogs, social media & Supports, Refutes, Neutral & \xmark &  Scientific Articles on COVID-19 & Coronavirus may have originated in bats or pangolins\\
        \midrule
        \sc covid-fact (2021) & COVID-19 & Reddit & Supported, Refuted & \xmark &  Google Search Results & Baricitinib restrains the immune dysregulation
in COVID-19 patients \\
        \midrule
        \sc covert (2022) & COVID-19 & Twitter & Supports, Refutes, Not Enough Information & \xmark & Google Search Results & 5G networks caused covid \\
\midrule
        \sc redhot (2023) & Medical Conditions & Reddit & \textit{N/A} & \textit{N/A} & Randomized Controlled Trial Abstract & Link between RA and migraines \\
        \midrule
                \sc healthfc (2024) & Health & Medizin Transparent &  Supported, Refuted, Not Enough Information & \checkmark & Systematic Review and Clinical Trial & Does cat’s claw improve joint disease symptoms?\\
                \midrule \rowcolor{green!10}
                        \sc our case study (2025) & Health & Reddit &  No Relevant Abstracts, Refutes, Partially Refutes, Inconclusive, Partially Supports, Supports & \checkmark & Randomized Controlled Trial Abstract & \textbf{Contextualized Claim} (Table~\ref{table:grapefruit})\\

            \bottomrule
    \end{tabular}
    \caption{Comparative overview of related work in medical and health fact-checking. \textit{N/A} indicates components not applicable to the task (e.g., \textsc{RedHOT} does not perform claim verification).}
    \label{table:relatedwork}
\end{table*}

This position paper highlights practical gaps in AI-driven fact-checking systems that cannot be addressed by ``building a better system'' alone.
We focus on real-world medical claims from social media, preserving their original context, and
contend that fact-checking systems should mirror how experts (e.g., physicians, care providers) evaluate and respond to such claims. 
As part of our exploration, we designed an annotation study that examines how medical experts verify claims using retrieved medical evidence.
Experts were asked to assess medical claims present on social media (i.e., Reddit forums about a particular medical condition) by synthesizing retrieved randomized controlled trial (RCT) abstracts and explaining their judgments. 
This provides an idealized upper bound for systems verifying \textit{``in the wild''} claims. 
However, we highlight fundamental obstacles that challenge the construct validity of end-to-end automated systems for fact-checking: given a claim, provide a veracity judgment.
We identify inherent difficulties in this setup even for domain experts, including: 
connecting claims with evidence; ambiguity from underspecified claims leading to valid yet contradictory interpretations, a challenge also identified in prior work on evidence-based fact-checking \citep{glockner-etal-2024-ambifc}; and challenges in achieving annotator consensus due to the inherent subjectivity of veracity labels. 
These issues suggest that the existing framing of fact-checking as an end-to-end classification task is inadequate for real-world settings, which may explain in part why such systems have not been put into wide use. 

To correct the flawed construct validity of this task, we contend that \textbf{fact-checking should be an interactive dialogue agent rather than an end-to-end system}. 
We envision a human-centered \textbf{communication model} for medical fact-checking inspired by interactions between patients and physicians. We explain how this model can overcome existing challenges and empower experts and laypeople to engage in constructive medical discourse.

\section{Background: Medical Claim-Checking}

\label{previous}

\citet{guo-etal-2022-survey} outlines the conventional  framework for automated fact-checking which comprises three stages: \textbf{(1) Claim Detection}, \textbf{(2) Evidence Retrieval}, and \textbf{(3) Claim Verification}. 
In the \textbf{Claim Detection} stage, the system identifies claims—statements asserting verifiable facts—and often ranks them based on check-worthiness factors such as public interest, popularity, timeliness, and impact \citep{DAS2023103219, micallef2022true}. Complex claims may also be automatically decomposed into sub-claims for individual verification \citep{wanner-etal-2024-closer, pan-etal-2023-fact, min-etal-2023-factscore, kamoi-etal-2023-wice, jing-etal-2024-faithscore}. 
\textbf{Evidence Retrieval} entails retrieving supporting evidence to inform verification  \citep{chen2024complexclaimverificationevidence}. Finally, \textbf{Claim Verification} requires determining the claim's veracity and generating a justification grounded in the retrieved evidence. There is a growing interest in using LLMs to automate these stages of the fact-checking pipeline \citep{vykopal2024generativelargelanguagemodels, iqbal-etal-2024-openfactcheck, 10.3389/frai.2024.1341697}.

\begin{figure*}[t]
\centering
\includegraphics[width=\textwidth]{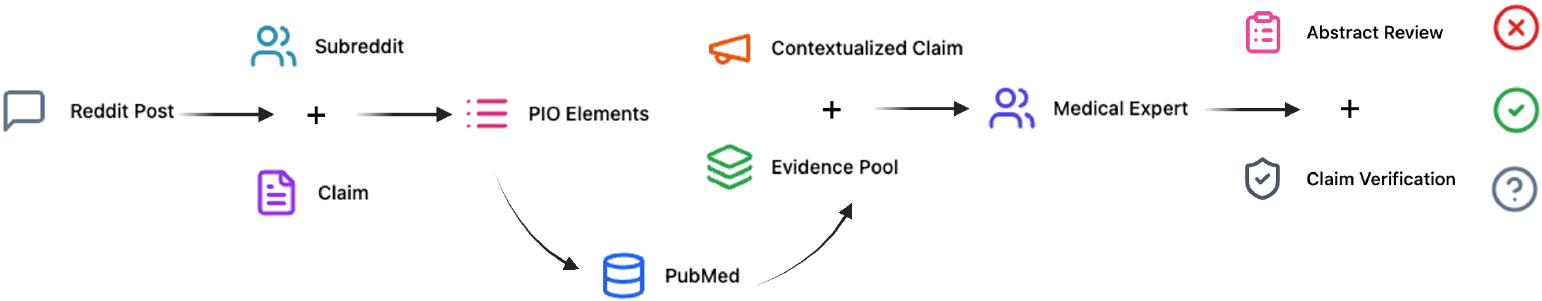}
\caption{Overview of our AI-in-the-loop expert study pipeline. Given a claim from a subreddit, we extract PIO elements and automatically retrieve evidence. The evidence and its context are presented to a medical expert, who provides a veracity judgment and grounded rationale.}
\label{fig:task}
\end{figure*}

We present a comparative overview of prior work in medical fact-checking in Table~\ref{table:relatedwork}. With few exceptions (e.g., \cite{wadhwa-etal-2023-redhot} which we use in this study), existing work largely views claims as statements that ``stand alone'' without context, and has treated fact-checking as an end-to-end pipeline with the last step as a multi-label classification task \citep{sarrouti-etal-2021-evidence-based}.
Additionally, while prior work has examined expert assessment of the credibility of web-based medical information \citep{info:doi/10.2196/26065}, no work has yet examined \textbf{expert involvement in every stage of medical claim checking}---a fine-grained examination of claim interpretation, retrieved evidence, and veracity judgment, with natural language explanations. As a result, 
we do not yet have an understanding of an upper-bound that modern systems could achieve on evidence-based medical claim verification. Finally, while formal evidence synthesis has long been studied in the health literature \citep{10.1093/asj/sjv130, Sackett71, Moberg2018-dx, Cumpston19}, it has not been integrated into LLM-based medical fact-checking systems.

\section{Case Study Methods} 
To reveal the gaps in end-to-end fact-checking systems, we set out to establish an ``idealized'' scenario: given the task construction for automatic fact-checking systems discussed in Section~\ref{previous}, we asked domain experts to perform all subtasks that are delegable to humans.
Thus, we formulate the following analytic study for expert medical claim checking using the AI-in-the-loop pipeline shown in Figure~\ref{fig:task}.

\paragraph{Claim source} We use claims from the Reddit Health Online Talk (RedHOT) corpus~\citep{wadhwa-etal-2023-redhot}, which contains 22,000 annotated posts from Reddit across 24 health conditions. 
RedHOT defines a claim as a statement indicating (often only implicitly) a causal relationship between an intervention and an outcome. 
To help experts contextualize these claims, we provide the full post and \textbf{P}opulation, \textbf{I}ntervention, \textbf{O}utcome (PIO) descriptors. (See Appendix~\ref{appendix:pioelements}, Appendix~\ref{appendix:pioextraction} for derivation details.) Note that we leave off the \textbf{C}omparators because in practice claims rarely mention the comparator explicitly (e.g., ``Vitamin C cured my flu'').
Given a contextualized medical claim, 
a medical expert
constructs a hierarchy of evidence \citep{Guyatt995, Guyatt924, GUYATT2011383} based on relevance and quality, assesses the claim's veracity, and provides a grounded explanation. 

\paragraph{Annotation task setup}
The overall task, without any aid, places a high cognitive load on experts who must comb through and synthesize multiple pieces of evidence~\citep{10.1145/3555143}. With the aim of easing this, 
we automated several steps of the fact-checking process and integrated them into an intuitive web-based annotation interface (see Appendix~\ref{appendix:annointerface}).  
These features enable experts to focus on critical aspects: evaluating evidence relevance and synthesizing it to support or refute claims. 
Detailed annotation guidelines are provided in Appendix~\ref{appendix:annoguidelines}.

For each claim, we used its PIO elements to automatically retrieve ten published relevant RCT abstracts from Trialstreamer ~\citep{10.1093/jamia/ocaa163}, a continuously updated database of RCTs, as potential evidence.
We provide experts with these RCT abstracts and their publication dates. We used a dense retrieval system using state-of-the-art embedding models. We provided a more detailed description of this retrieval methodology in Appendix~\ref{appendix:retrieval}.
We do not evaluate the feasibility of manual evidence search in this study. However, experts provide judgments on how relevant retrieved abstracts are to the claim in question.
For each RCT abstract, we collected annotations determining its \textbf{relevance} to the claim along four dimensions: Population, Intervention, Outcome (PIO), and overall relevance. 
Claims are \emph{contextualized} in their original Reddit posts during annotation. 
Annotators labeled each dimension as \textbf{(1)} Relevant, \textbf{(2)} Somewhat Relevant, or \textbf{(3)} Irrelevant.
If an abstract was deemed \textit{overall} relevant, annotators highlighted the most relevant text span and assessed whether the trial described in the abstract supports or refutes the claim using the four labels: \textbf{(1)} Supports, \textbf{(2)} Partially Supports, \textbf{(3)} Partially Refutes, and \textbf{(4)} Refutes. 

After annotating all ten abstracts, experts proceeded to synthesize the evidence. 
To help experts navigate through the evidence documents, we implemented a tiering step. 
Abstracts are initially tiered automatically based on their relevance annotations in the previous step, establishing a natural hierarchy. Annotators are free to further refine this hierarchy by considering evidence quality. Next, annotators verify the claim in two phases:
\setlength{\leftmargini}{16pt}
\setlength{\itemsep}{0pt}
\setlength{\parsep}{0pt}
\setlength{\topsep}{0pt}
\begin{enumerate}
    \item \textbf{Overall Support:} Verification based solely on the provided evidence.
    \item \textbf{Expert Support:} Optional verification based on clinical expertise, particularly for claims unlikely to be studied in clinical trials.
\end{enumerate}
This separation allows us to compare expert opinion with evidence-based conclusions and minimize bias. Annotators select from six labels for both phases: \textbf{(1)} No Relevant Abstracts/No Expert Opinion (for each of the above phases respectively), \textbf{(2)} Refutes, \textbf{(3)} Partially Refutes, \textbf{(4)} Inconclusive, \textbf{(5)} Partially Supports, and \textbf{(6)} Supports. 

To justify their veracity label, experts write a paragraph-length explanation (see guideline in Appendix~\ref{appendix:plainlanguageexplanationformat}). Annotators may optionally include a medical addendum detailing clinical practices typically used in response to the claim, providing practical context for users.

\paragraph{Annotation team}
Our annotation team consists of five clinical experts, one serving as the medical lead. All experts had experience reviewing medical articles and synthesizing them for biomedical research or patient care (annotator recruitment details are in Appendix~\ref{appendix:annorecruitment}). 
To leverage their expertise effectively, we conducted the study in \textbf{three} rounds, with changes between rounds detailed in Appendix~\ref{appendix:roundschanges}. Following ~\citet{10.1162/coli_a_00516}, the two co-first authors held group meetings with the experts during each round to discuss disagreements and refine the annotation guidelines. In total, these meetings spanned four hours.

Across the three rounds, each of the five experts annotated 20 unique claims, yielding 1,000 abstract-level annotation instances (10 abstracts per claim) and 100 synthesis-level explanations.

\section{Challenges for End-to-End Fact-Checking}
\label{sec:analysis}

In this section, we present the fundamental challenges revealed by our expert annotations and inputs, gathered over multiple rounds of discussion.
\subsection{Challenge 1: Connecting Medical Evidence with Claims}
\label{sec:evidenceclaim}

We present inter-annotator agreement from the final round of annotations for five claims, each annotated by five experts (totaling 25 separate verifications) in Table~\ref{table:finalanno}. This round includes 50 abstracts per expert (250 total), with five labels per abstract and two labels per claim.
Despite multiple rounds of expert feedback to improve the annotation task, agreement remained low across all fields. 
A common interpretation of $\kappa$ scores is that values of 0.21--0.40 indicate fair agreement, 0.41--0.60 moderate agreement, and 0.61--0.80 substantial agreement \citep{Landis1977TheMO}. Because agreement is generally harder to achieve in multi-class labeling settings \citep{10.1093/ptj/85.3.257}, we consider a $\kappa$ score of approximately 0.5 to be a reasonable target in our setting.
These results should not be overgeneralized. Rather, they describe substantial variability in expert judgments under shared evidence conditions and the inability of experts to converge when evaluating the same claims using the same evidence, calling into question the construct validity of this task. 

\begin{table}
\small 
    \centering
    \begin{tabular}{lc}
    \toprule
    \textbf{Type} & \textbf{$\kappa$ ($\uparrow$)}\\  \midrule
        \textcolor{cyan}{Population} & 0.416\\
        \textcolor{cyan}{Intervention} & 0.714  \\
        \textcolor{cyan}{Outcome} & 0.200 \\
        \textcolor{cyan}{Overall (Abstract-level)} & 0.155 \\
        \textcolor{cyan}{Tab Support} & 0.170 \\
        \textcolor{magenta}{Overall Support} & 0.124\\
        \textcolor{magenta}{Expert Support} & -0.184\\
        \bottomrule
    \end{tabular}
    \caption{Inter-annotator agreement for each component of the fact-checking pipeline in the final round of five claims. \textcolor{cyan}{Blue}: abstract-level labels; \textcolor{magenta}{pink}: synthesis-level labels.}
    \label{table:finalanno}
\end{table}

\begin{table*}[t]
    \centering
    \scriptsize
\begin{tabular}{
p{0.97\textwidth}} \toprule
        \RHExample{Grapefruit}{Grapefruit and siezures
        
        Guys, I am wondering if you have any issues or know about interactions between oxcarbazepine and/or levetiracetam and grapefruit? \hlyellow{I believe it may make those medications work differently, but I am not sure.}}{Epilepsy patients (implied by the subreddit r/Epilepsy and the mention of seizure medications)}{Grapefruit consumption (in interaction with oxcarbazepine and/or levetiracetam)}{Medication efficacy (i.e., how the medications work)}{r/Epilepsy} 
        \vspace{0.3em} \\
    \bottomrule
\end{tabular}
\vspace{-0.5em}
\caption{An example of an unverifiable claim, as no RCTs have examined interactions between grapefruit, oxcarbazepine, and epilepsy, and such a study may be infeasible.}
\label{table:grapefruit}
\end{table*}

Most instances---20 out of the 25 expert judgments---were labeled ``No Relevant Abstracts'', indicating that the claims were unverifiable.
For three of the five claims in this final round, all experts independently labeled them ``No Relevant Abstracts'' (see Section~\ref{sec:trikafta_example} for an illustrative example). This high rate of unverifiable claims underscores a broader challenge: even when claims are annotated to suggest a causal relationship between an Intervention and an Outcome \citep{wadhwa-etal-2023-redhot} and are paired with state-of-the-art evidence retrieval, they are often unverifiable in practice.

Our expert annotation study identifies four reasons why:
\setlength{\leftmargini}{16pt}
\begin{enumerate}
    \item \textbf{No Intervention:} Claims lacking an intervention cannot be verified through an RCT.
    \item \textbf{Unethical Intervention:} Some interventions are unethical to test via RCTs because they may harm participants. For example, it is unethical to study smoking as an intervention in an RCT.
    \item \textbf{Lack of Feasibility:} Claims involving specific PIO element combinations are often unverifiable due to the impracticality or improbability of conducting such RCTs.
    \item \textbf{Lack of Utility:} Some claims, while theoretically verifiable through an RCT, lack available evidence as findings from such studies would lack utility in the medical field.
\end{enumerate}

These issues highlight the difficulty of collecting evidence that is directly relevant to claims people make (on social media).
Medical evidence, especially high-quality evidence, is constrained by standards for feasibility and ethics, making it impossible to cover all possible queries from the public.
Prior to annotation, we used an automated method to filter out claims that could not be verified by RCTs (detailed in Appendix~\ref{appendix:rctpipeline}). 
This issue of unverifiability remained despite this effort.
Part of addressing this challenge lies in expanding the pool of evidence while ensuring it remains trustworthy (see Section~\ref{sec:rct_solutions}).
Additionally, systems should tackle how to best handle the inevitable case in which a claim is unverifiable. 
We discuss a guided retrieval approach to this in Section~\ref{sec:guided_retrieval}.

\newcommand{\Example}[7]{
    \textbf{#1} \vspace{0.3em}
    \newline
    \textbf{Post (#7):} 
    #2 
    \vspace{0.3em}
    \newline
    \textbf{Population:} \fcolorbox{black}{orange!20}{#3}
    \textbf{Intervention:} \fcolorbox{black}{magenta!20}{#4}
    \textbf{Outcome:} \fcolorbox{black}{cyan!20}{#5}
    \vspace{0.3em}
    \newline
    \textbf{Expert Feedback:} \vspace{0.3em} \newline
    #6
}
\begin{table*}[t]
    \centering
    \scriptsize
\begin{tabular}{
p{0.97\textwidth}} \toprule
        \Example{ADHD, Herbs, and Menstruation}{Hello, menstruating people! How do your cycle and ADHD influence each other and how do you deal with it?

EDIT: After getting your responses I am reflecting again how medicine does not give a shit about women. It's truly insane. Thank you!

Hello! I have never paid too much attention to my menstrual cycle since it was never particularly bothersome. Now that I take methylo I feel big changes in how I function during the cycle. Like last 10 days of the cycle, my medication kind of stops working... That is like 1/3 of the time. I know it's still better than without meds nevertheless, it makes establishing a routine quite challenging. My doc suggested trying contraceptive pills, but I am not even sexually active ATM so taking more medication, with potential side effects, does not excite me.

\hlyellow{I know there are herbs that are proven to be helping with regulating the cycle} but I don't know if they would help with ADHD symptoms? Any tips?}{People with ADHD}{Herbs}{Regulating the menstrual cycle}{
            - \textit{What does regulating the cycle mean?} \newline
            - \textit{ADHD has no bearing on one's menstruation cycle. It is a red herring.} \newline
            - \textit{Trials with these descriptors are unlikely to exist.} \newline
            - \textit{Is this claim really what the patient is concerned about in this post?} 
        }{r/ADHD} 
        \vspace{0.3em} \\
        \rowcolor{lightgray!20}
        \Example{Pineapple Juice Reduces Inflammation}{
        Anyone with sinus issues drinking pineapple juice?

It's a weird question, but \hl{I saw a post about pineapple juice being good for sinus issues (helps with the inflammation)} and just wondered if anyone has done this? Some people were commenting about the high sugar content in pineapple juice not being good, but they get around that by taking a supplement instead of drinking the juice. Anyone?}{Patients With Cystic Fibrosis}{Pineapple Juice}{Reduced Inflammation/Fewer Sinus Issues}{
    - \textit{How quickly is the poster expecting the intervention to produce results?} \newline
    - \textit{Just improving inflammation should not be the only criteria.} \newline
    - \textit{Trials with these exact descriptors are unlikely to exist.} 
}{r/CysticFibrosis} \\
    \bottomrule
\end{tabular}
\caption{Examples of underspecified claims and corresponding expert feedback.}
\vspace{-5pt}
\label{table:chal2_ex}
\end{table*}

\subsection{Challenge 2: Variations in the Interpretation of Claims}
\label{sec:chal2}

In our discussions with annotators, we noted consistent disagreements in how they interpreted medical claims, which in turn caused disagreement in the claim verification task.
To address this, we provided annotators with PIO descriptors (Table~\ref{table:grapefruit}) of claims to narrow the scope of possible interpretations. 
However, even with this added context reaching consensus among annotators remained a challenge.
This is because \textit{claims in the wild} about health tend to be \textit{underspecified} and/or \textit{misguided}, causing annotators to deduce their own varying interpretations on what the \textit{\claimant}, the author of the claim (usually a layperson), intended.

\paragraph{Underspecified Claims}
Naturally occurring medical claims on social media are usually written informally by laypeople, and tend therefore to be underspecified (see Table~\ref{table:chal2_ex}). 
For example, a patient with ADHD claimed that ``herbs are proven to be helping with regulating the cycle.'' 
It is unclear what ``regulating'' means here, and annotators interpreted this in various ways, e.g., reducing symptoms related to menstruation or skipping menstruation altogether. 
Another \claimant{} claimed that pineapple juice is ``good for sinus issues'' by reducing inflammation. 
However, it was unclear whether they meant immediate or gradual improvement (no time frame was offered). 
Resolving these underspecifications requires understanding the \claimant{}s' intentions, which is challenging since intent cannot always be inferred from the claim and its context alone.

\paragraph{Misguided Claims}
Discussions with annotators also identified another artifact of naturally occurring medical claims: \textit{misguided} claims formed from incorrect premises. 
Annotators often disagreed on how to handle such claims within our task.
The previously mentioned ADHD example illustrates this issue. 
The patient, prescribed methylphenidate, noticed no effect during the last 10 days of their menstrual cycle. Their doctor suggested contraceptive pills as a potential solution. Concerned about the side effects of contraceptives, the patient considered using herbs as an alternative to ``regulate the cycle''. 
Annotators characterized the underlying premise---that the medication’s efficacy is affected by the menstrual cycle---as incorrect. 
Disagreement over whether to consider the premise's validity in veracity judgments led to conflicting assessments among annotators.
Prior work in general-domain fact-checking used claim decomposition to address false presuppositions in claims \citep{chen-etal-2022-generating, kamoi2023wicerealworldentailmentclaims, hu2025decompositiondilemmasdoesclaim}, however this does not tackle \emph{implicit} premises.

\paragraph{Mismatched Intent}
Previous work on fact-checking has addressed underspecified claims, often by decontextualizing them, i.e., removing context and resolving underspecifications based on local content~\citep{deng2024documentlevelclaimextractiondecontextualisation, gunjal-durrett-2024-molecular}. 
This approach can clarify underspecifications, but it disconnects the claim from the \claimant's original intent, as embedded in the global context. This disconnect can result in verifications that do not apply to the original claim, allowing subtle falsehoods to slip through and potentially be amplified. 
For example, suppose the underspecification of ``regulating the cycle'' were resolved and the claim were deemed true without considering its premise, 
this verification would fail to address the \claimant's true goal of improving the efficacy of their ADHD medication. 
Such verification would also implicitly validate the misunderstood premise. 
It is also the case that the \claimant's true intention is not about this particular claim, but an overall desire to communicate and discuss the underlying condition to get better.
To effectively address this, the focus must be on meeting the \claimant's 
\textit{information needs}, and when needed, uncovering and assessing their assumptions.

\begin{figure*}[t]
\centering
\vspace{-10pt}
\includegraphics[width=\textwidth]{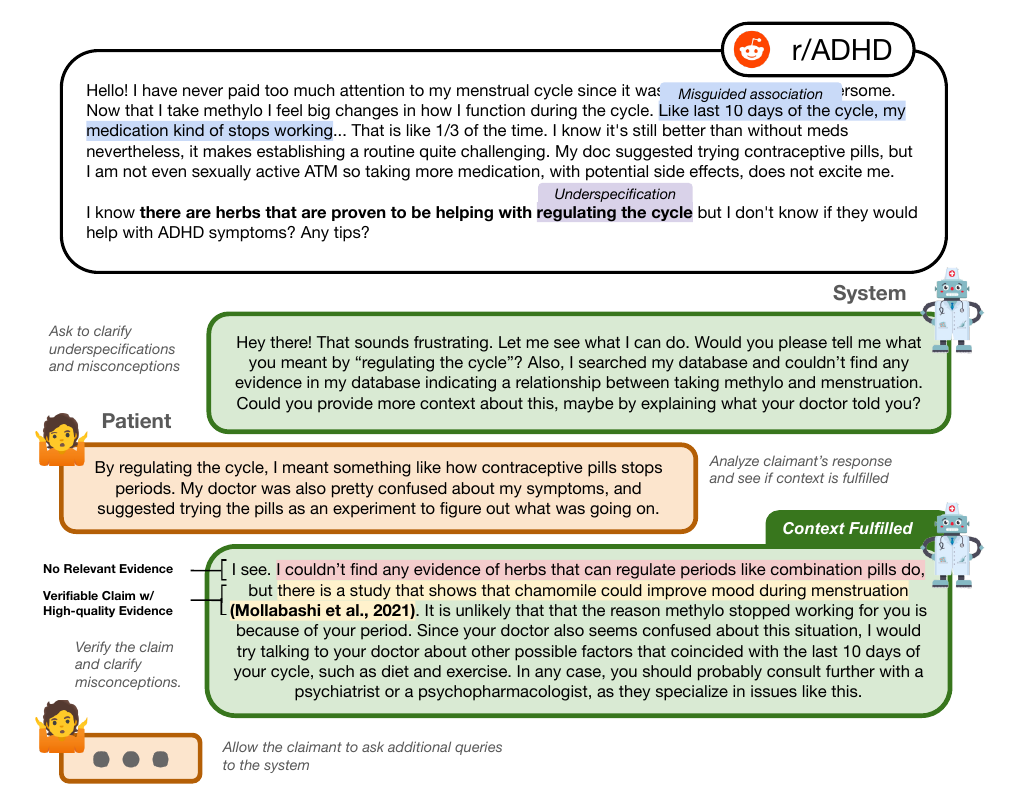}
\vspace{-20pt}
\caption{Communication model for fact-checking, where the system engages the patient by asking clarifying questions, filling contextual gaps, and verifying claims while addressing misconceptions.}
\vspace{-5pt}

\label{fig:communication}
\end{figure*}

\subsection{Challenge 3: Labeling the Severity of Inaccurate Statements is Inherently Subjective}
\label{sec:veracity_label}

Another factor that contributed to the disagreements we observed is the subjectivity in labeling the \textit{veracity}, or the degree of truth of a medical claim. 
This subjectivity, as we observe, is not caused by differing interpretations of the claim.
Rather, experts are influenced by their backgrounds and philosophies, applying different standards for assessing a claim's ``truthfulness'' based on evidence.

Consider the claim about pineapple juice in Table~\ref{table:chal2_ex}. 
If the desired onset for resolving sinus issues is clarified to mean within a few days, the claim is technically false. 
However, the \emph{severity} of this falsehood is subjective. One expert might view it as a minor inaccuracy, noting that pineapple juice may help with sinus issues but works more slowly and less effectively than targeted treatments. Another expert might see it as a serious falsehood, arguing that promoting pineapple juice as a quick fix is misleading and potentially harmful. Both perspectives are valid, highlighting the inherent subjectivity in judging a claim's truthfulness. Experts’ sensitivities can also vary depending on the topic, leading to apparently inconsistent judgments. 
These findings suggest that the existing practice---simply asking for a veracity label---needs to be redefined to align expert opinions in the first place.

\section{How Can We Address These Challenges? A Communication Model}
To address the challenges in Section~\ref{sec:analysis}, we argue that fact-checking alone is insufficient: a \textbf{communication model} is required to mirror a dialogue between a healthcare provider and a patient. This frames fact-checking as a dialogue aimed at addressing the \claimant's information needs.
A vision of the user interaction with the communication model is shown in Figure~\ref{fig:communication}. 
\subsection{Clarifying Intent Through Conversation}
\label{future:intent}
In Section~\ref{sec:chal2}, we describe how naturally occurring claims often contain underspecifications that require understanding the \claimant's intentions. A communication model can address this through dialogue with the \claimant{}, similar to clinical interactions where physicians ask patients questions to understand their care needs. Prior work in dialogue systems has explored resolving ambiguity by generating clarification questions and modeling future conversations ~\citep{kim2023treeclarificationsansweringambiguous,zhang2024modelingfutureconversationturns, zhang2023clarifynecessaryresolvingambiguity}. 

To do this effectively, the system must identify underspecifications from the provided context.
Similar work identifying ambiguities in user queries could provide a template for this task~\citep{zhang2024clamberbenchmarkidentifyingclarifying}.
However, as we discovered in our analysis, identifying these underspecifications also requires extensive expert knowledge of the claim's subject matter, which could be encoded in a trained model or accessed via a retrieval-augmented system.
The system must also identify ``misguided'' claims, as discussed in Section~\ref{sec:chal2}, which requires recognizing subtextual implications and commonly held misconceptions.
An ideal system would proactively query the \claimant{} to uncover incorrect assumptions and address them with empathy, fostering constructive engagement.

Direct communication with the {\claimant}  
is not strictly necessary to achieve intent clarification. 
\cite{kim2023treeclarificationsansweringambiguous} proposed generating a tree of clarification questions, which, when fully answered, provides the context needed to resolve an ambiguous query. 
A similar approach could be applied here, where a tree of clarification questions resolves different interpretations of an underspecified claim, all of which must be verified and included in the system's final output. 
However, for this approach to work research is needed to study how to align the final output with the \claimant's original intent.

\subsection{Guided Retrieval of Medical Evidence}
\label{sec:guided_retrieval}
In Section~\ref{sec:evidenceclaim}, we discussed the disconnect between what is practical and measurable in evidence-based medicine and what patients care about. The communication model enables providers to guide such claims toward verifiability while clarifying the \claimant's intent. 
To support this process, evidence retrieval should inform the dialogue between the \claimant{} and the system. 
When no relevant evidence is found, the system should communicate this and guide the \claimant{} toward related, verifiable claims. 
When none is available, the system should \emph{abstain}. This approach makes clear that the claim is unverifiable while offering a pathway for continued learning. 

This \textit{guided retrieval} could also help correct misguided claims, as this conversational approach digs into and exposes the \claimant's thought process. 
Similar to physician-patient interactions, this process resembles a physician's response to an unanswerable query.
They might first gather more information about the patient's question; if it remains unanswerable, they might recall related (answerable) questions.
The field of Interactive Information Retrieval (IIR) studies the modeling and optimization of  such back-and-forth interactions between users and retrieval systems~\citep{zhai-2020-interactive}, e.g., for product retrieval~\citep{wang2024interactivemultimodalqueryanswering, aliannejadi2024interactionsgenerativeinformationretrieval}.
Similarly, this approach could be used to guide unverifiable claims---often misguided due to false assumptions---toward claims that satisfy the \claimant's \textit{information needs}. 

\subsection{Communicating Veracity Through Diverse Perspectives}
As discussed in Section~\ref{sec:veracity_label}, our annotation study demonstrated that categorizing claims with fine-grained veracity labels is inherently subjective. 
While annotators often disagreed on labels, their reasoning in plain language explanations was often similar. 
During discussions, annotators often accepted each other's explanations as valid despite disagreeing on the level of severity. 
This suggests that a wider range of ostensible ``agreement'' can be reached. 
We propose that an effective medical fact-checking agent should produce responses that reflect diverse expert perspectives, acknowledging the inherent heterogeneity of expert evaluation.
The need for these diverse explanations is corroborated by fact-checking professionals, who acknowledge the need and complexity of crafting thorough, nuanced explanations and calls for explanations to accommodate different audience needs~\citep{10.1145/3706598.3713277}.
Encouraging response diversity, rather than imposing artificial consensus via a single numerical value, is crucial for developing multi-agent medical fact-checking systems that integrate multiple expert viewpoints. 

\subsection{Medical Evidence Beyond RCTs}
\label{sec:rct_solutions}
To bridge the gap between user questions and the limited pool of RCTs, future work could incorporate other forms of medical evidence such as meta analyses, cohort studies, case-control studies, case series, and case reports. Expanding the scope of medical evidence increases complexity, as systems must determine how to retrieve and synthesize evidence of various types and strengths, how to appropriately communicate this to end users. 
To this end, systems should adequately communicate its uncertainty according to existing guidelines in medicine~\citep{ratcliff2021impact, simpkin2019communicating} and reference lower-grade evidence only when the communication model fails to identify a helpful, verifiable question and should clearly convey the quality of the source evidence and its limitations.

\subsection{Multi-Party Applications}

This communicative approach to fact-checking opens various avenues for applications. In addition to the one-on-one communication described in Figure~\ref{fig:communication}, one could also imagine a one-on-many approach, in which a system interacts within a forum of laypeople. Such an application could easily be inserted in platforms like Reddit to counter the spread of false claims, and would function as a mediated discussion around health claims with an LLM in the loop.  

\section{Conclusion}
\label{sec:conclusion}
In this position paper, we present an analysis of the construct validity of existing end-to-end automatic medical fact-checking systems, with expert engagement in all key aspects of the system. Our work highlights the unique challenges of automated medical fact-checking, showing that it should be approached with user interaction in mind, and not as an end-to-end system. We propose a communication model to clarify underspecifications and guide unverifiable claims, aiming to improve user outcomes and real-world utility. We hope this work inspires further exploration of human-in-the-loop systems for medical fact-checking. 

\section*{Limitations} 
\label{sec:limits}
Due to time and cost constraints, our annotation study includes a limited number of claims. However, each claim was evaluated against 10 retrieved abstracts, resulting in a substantial volume of evidence-based annotations. These annotations and expert discussions revealed fundamental challenges in medical fact-checking that make large-scale annotation difficult to define in a principled way.

We used an automatic document retrieval system instead of manual search by annotators to identify relevant abstracts for each claim. While this approach may fail to retrieve the most relevant evidence, retrieval performance was not the focus of this study. Rather, our goal was to examine how retrieved medical evidence is synthesized to verify claims, while avoiding undue burden on expert annotators.

These limitations reflect broader challenges in defining and evaluating medical fact-checking tasks in practice. As for the proposed communication model, the main limitation lies in potential barriers to speed and scalability. However, we consider this to be a worthy tradeoff for the deeper and nuanced clarifications this approach would elicit, which is especially necessary in fact-based discussions. We leave the solutions for these limitations up to future work. 

\section*{Ethical Considerations}
\label{app:ethics}

The posts found within the RedHOT dataset contain health-related comments that are inherently sensitive. 
To respect this sensitivity, the authors of the RedHOT dataset notified all users of their inclusion in this dataset and provided them with the opportunity to opt-out.
They also did not release the data directly, but instead provided a script to download content from Reddit so that individuals can remove their post in the future.
In this work, we directly release a small subset of these posts from RedHOT that we used in our annotation study.
In our released data, we do not reveal the username of the author of the post. We only include the text from the post and information about the subreddit in which it was found.
Considering the measures taken by the authors of the RedHOT dataset and the fact that these posts have been publicly available on Reddit for at least a year, we believe it is safe to publicly release this data.

We have consulted with our Institutional Review Board (IRB) about the nature of our work and confirmed that the use of the RedHOT data and the subsequent annotation study using these data do not constitute research of human subjects. 
However, we acknowledge that certain uses of this data may be considered sensitive. 
We strongly encourage researchers to obtain prior approval from their own IRB regarding the intended use of the data released by this work.

We also consider ethical aspects of the annotation process. Annotators consented to participate and were compensated at a competitive hourly rate (see Appendix~\ref{appendix:annorecruitment} for details).

\section*{Acknowledgments}
We thank our medical experts Barry Wei, Gustavo Antonio Lima de Campos, Jon Stewart Hao Dy, Beatrix Krause-Sorio, Aislin Azcárate Kent, and Celina Bandowski for their annotation efforts. This work is partially supported by Good Systems\footnote{\url{https://goodsystems.utexas.edu}}, a UT Austin Grand Challenge to develop responsible AI technologies, and National Institutes of Health (NIH) grant 1R01LM014600-01. 
We also acknowledge support from the National Science Foundation (NSF) grant 2211954 and the Wellcome Trust grant 313618/Z/24/Z. 

Lily Chen was supported by the DOE CSGF. This material is based upon work partially supported by the U.S. Department of
Energy, Office of Science, Office of Advanced Scientific Computing Research, Department of Energy Computational Science Graduate Fellowship under Award Number DE-SC0026073. This report was prepared as an account of work sponsored by an agency of the United States Government. Neither the United States Government nor any agency thereof, nor any of their employees, makes any warranty, express or implied, or assumes any legal liability
or responsibility for the accuracy, completeness, or usefulness of any information, apparatus,
product, or process disclosed, or represents that its use would not infringe privately owned
rights. Reference herein to any specific commercial product, process, or service by trade name,
trademark, manufacturer, or otherwise does not necessarily constitute or imply its
endorsement, recommendation, or favoring by the United States Government or any agency
thereof. The views and opinions of authors expressed herein do not necessarily state or reflect
those of the United States Government or any agency thereof.
\bibliography{custom}

\begin{thebibliography}{52}
\providecommand{\natexlab}[1]{#1}

\bibitem[{Aliannejadi et~al.(2024)Aliannejadi, Gwizdka, and Zamani}]{aliannejadi2024interactionsgenerativeinformationretrieval}
Mohammad Aliannejadi, Jacek Gwizdka, and Hamed Zamani. 2024.
\newblock \href {https://arxiv.org/abs/2407.11605} {Interactions with generative information retrieval systems}.
\newblock \emph{Preprint}, arXiv:2407.11605.

\bibitem[{Bastian et~al.(2010)Bastian, Glasziou, and Chalmers}]{bastian2010seventy}
Hilda Bastian, Paul Glasziou, and Iain Chalmers. 2010.
\newblock Seventy-five trials and eleven systematic reviews a day: how will we ever keep up?
\newblock \emph{PLoS medicine}, 7(9):e1000326.

\bibitem[{Champlin et~al.(2017)Champlin, Mackert, Glowacki, and Donovan}]{champlin2017toward}
Sara Champlin, Michael Mackert, Elizabeth~M Glowacki, and Erin~E Donovan. 2017.
\newblock Toward a better understanding of patient health literacy: A focus on the skills patients need to find health information.
\newblock \emph{Qualitative Health Research}, 27(8):1160--1176.

\bibitem[{Chen et~al.(2024)Chen, Kim, Sriram, Durrett, and Choi}]{chen2024complexclaimverificationevidence}
Jifan Chen, Grace Kim, Aniruddh Sriram, Greg Durrett, and Eunsol Choi. 2024.
\newblock \href {https://arxiv.org/abs/2305.11859} {Complex claim verification with evidence retrieved in the wild}.
\newblock \emph{Preprint}, arXiv:2305.11859.

\bibitem[{Chen et~al.(2022)Chen, Sriram, Choi, and Durrett}]{chen-etal-2022-generating}
Jifan Chen, Aniruddh Sriram, Eunsol Choi, and Greg Durrett. 2022.
\newblock \href {https://doi.org/10.18653/v1/2022.emnlp-main.229} {Generating literal and implied subquestions to fact-check complex claims}.
\newblock In \emph{Proceedings of the 2022 Conference on Empirical Methods in Natural Language Processing}, pages 3495--3516, Abu Dhabi, United Arab Emirates. Association for Computational Linguistics.

\bibitem[{Cumpston and Thomas(2019)}]{Cumpston19}
Li~T Page MJ Chandler J Welch VA Higgins~JPT Cumpston, M and J~Thomas. 2019.
\newblock \href {https://doi.org/10.1002/14651858.ED000142} {Updated guidance for trusted systematic reviews: a new edition of the cochrane handbook for systematic reviews of interventions}.
\newblock \emph{Cochrane Database of Systematic Reviews}, (10).

\bibitem[{Das et~al.(2023)Das, Liu, Kovatchev, and Lease}]{DAS2023103219}
Anubrata Das, Houjiang Liu, Venelin Kovatchev, and Matthew Lease. 2023.
\newblock \href {https://doi.org/10.1016/j.ipm.2022.103219} {The state of human-centered nlp technology for fact-checking}.
\newblock \emph{Information Processing \& Management}, 60(2):103219.

\bibitem[{Deng et~al.(2024)Deng, Schlichtkrull, and Vlachos}]{deng2024documentlevelclaimextractiondecontextualisation}
Zhenyun Deng, Michael Schlichtkrull, and Andreas Vlachos. 2024.
\newblock \href {https://arxiv.org/abs/2406.03239} {Document-level claim extraction and decontextualisation for fact-checking}.
\newblock \emph{Preprint}, arXiv:2406.03239.

\bibitem[{Fox and Duggan(2013)}]{fox2013health}
Susannah Fox and Maeve Duggan. 2013.
\newblock Health online 2013. pew research center.
\newblock \emph{National survey by the Pew Research Center’s Internet and American Life Project}.

\bibitem[{Glockner et~al.(2024)Glockner, Stali{\={u}}nait{\.{e}}, Thorne, Vallejo, Vlachos, and Gurevych}]{glockner-etal-2024-ambifc}
Max Glockner, Ieva Stali{\={u}}nait{\.{e}}, James Thorne, Gisela Vallejo, Andreas Vlachos, and Iryna Gurevych. 2024.
\newblock \href {https://doi.org/10.1162/tacl_a_00629} {{A}mbi{FC}: Fact-checking ambiguous claims with evidence}.
\newblock \emph{Transactions of the Association for Computational Linguistics}, 12:1--18.

\bibitem[{Gunjal and Durrett(2024)}]{gunjal-durrett-2024-molecular}
Anisha Gunjal and Greg Durrett. 2024.
\newblock \href {https://doi.org/10.18653/v1/2024.findings-emnlp.215} {Molecular facts: Desiderata for decontextualization in {LLM} fact verification}.
\newblock In \emph{Findings of the Association for Computational Linguistics: EMNLP 2024}, pages 3751--3768, Miami, Florida, USA. Association for Computational Linguistics.

\bibitem[{Guo et~al.(2022)Guo, Schlichtkrull, and Vlachos}]{guo-etal-2022-survey}
Zhijiang Guo, Michael Schlichtkrull, and Andreas Vlachos. 2022.
\newblock \href {https://doi.org/10.1162/tacl_a_00454} {A survey on automated fact-checking}.
\newblock \emph{Transactions of the Association for Computational Linguistics}, 10:178--206.

\bibitem[{Guyatt et~al.(2011)Guyatt, Oxman, Akl, Kunz, Vist, Brozek, Norris, Falck-Ytter, Glasziou, deBeer, Jaeschke, Rind, Meerpohl, Dahm, and Schünemann}]{GUYATT2011383}
Gordon Guyatt, Andrew~D. Oxman, Elie~A. Akl, Regina Kunz, Gunn Vist, Jan Brozek, Susan Norris, Yngve Falck-Ytter, Paul Glasziou, Hans deBeer, Roman Jaeschke, David Rind, Joerg Meerpohl, Philipp Dahm, and Holger~J. Schünemann. 2011.
\newblock \href {https://doi.org/10.1016/j.jclinepi.2010.04.026} {Grade guidelines: 1. introduction—grade evidence profiles and summary of findings tables}.
\newblock \emph{Journal of Clinical Epidemiology}, 64(4):383--394.

\bibitem[{Guyatt et~al.(2008{\natexlab{a}})Guyatt, Oxman, Kunz, Vist, Falck-Ytter, and Sch{\"u}nemann}]{Guyatt995}
Gordon~H Guyatt, Andrew~D Oxman, Regina Kunz, Gunn~E Vist, Yngve Falck-Ytter, and Holger~J Sch{\"u}nemann. 2008{\natexlab{a}}.
\newblock \href {https://doi.org/10.1136/bmj.39490.551019.BE} {What is {\textquotedblleft}quality of evidence{\textquotedblright} and why is it important to clinicians?}
\newblock \emph{BMJ}, 336(7651):995--998.

\bibitem[{Guyatt et~al.(2008{\natexlab{b}})Guyatt, Oxman, Vist, Kunz, Falck-Ytter, Alonso-Coello, and Sch{\"u}nemann}]{Guyatt924}
Gordon~H Guyatt, Andrew~D Oxman, Gunn~E Vist, Regina Kunz, Yngve Falck-Ytter, Pablo Alonso-Coello, and Holger~J Sch{\"u}nemann. 2008{\natexlab{b}}.
\newblock \href {https://doi.org/10.1136/bmj.39489.470347.AD} {Grade: an emerging consensus on rating quality of evidence and strength of recommendations}.
\newblock \emph{BMJ}, 336(7650):924--926.

\bibitem[{Hassan et~al.(2015)Hassan, Adair, Hamilton, Li, Tremayne, Yang, and Yu}]{Hassan2015TheQT}
Naeemul Hassan, Bill Adair, James~T. Hamilton, Chengkai Li, Mark Tremayne, Jun Yang, and Cong Yu. 2015.
\newblock \href {https://api.semanticscholar.org/CorpusID:79175} {The quest to automate fact-checking}.

\bibitem[{Hu et~al.(2025)Hu, Long, and Wang}]{hu2025decompositiondilemmasdoesclaim}
Qisheng Hu, Quanyu Long, and Wenya Wang. 2025.
\newblock \href {https://arxiv.org/abs/2411.02400} {Decomposition dilemmas: Does claim decomposition boost or burden fact-checking performance?}
\newblock \emph{Preprint}, arXiv:2411.02400.

\bibitem[{Iqbal et~al.(2024)Iqbal, Wang, Wang, Georgiev, Geng, Gurevych, and Nakov}]{iqbal-etal-2024-openfactcheck}
Hasan Iqbal, Yuxia Wang, Minghan Wang, Georgi~Nenkov Georgiev, Jiahui Geng, Iryna Gurevych, and Preslav Nakov. 2024.
\newblock \href {https://aclanthology.org/2024.emnlp-demo.23} {{O}pen{F}act{C}heck: A unified framework for factuality evaluation of {LLM}s}.
\newblock In \emph{Proceedings of the 2024 Conference on Empirical Methods in Natural Language Processing: System Demonstrations}, pages 219--229, Miami, Florida, USA. Association for Computational Linguistics.

\bibitem[{Jing et~al.(2024)Jing, Li, Chen, and Du}]{jing-etal-2024-faithscore}
Liqiang Jing, Ruosen Li, Yunmo Chen, and Xinya Du. 2024.
\newblock \href {https://doi.org/10.18653/v1/2024.findings-emnlp.290} {{F}aith{S}core: Fine-grained evaluations of hallucinations in large vision-language models}.
\newblock In \emph{Findings of the Association for Computational Linguistics: EMNLP 2024}, pages 5042--5063, Miami, Florida, USA. Association for Computational Linguistics.

\bibitem[{Juneja and Mitra(2022)}]{10.1145/3555143}
Prerna Juneja and Tanushree Mitra. 2022.
\newblock \href {https://doi.org/10.1145/3555143} {Human and technological infrastructures of fact-checking}.
\newblock \emph{Proc. ACM Hum.-Comput. Interact.}, 6(CSCW2).

\bibitem[{Kamoi et~al.(2023{\natexlab{a}})Kamoi, Goyal, Diego~Rodriguez, and Durrett}]{kamoi-etal-2023-wice}
Ryo Kamoi, Tanya Goyal, Juan Diego~Rodriguez, and Greg Durrett. 2023{\natexlab{a}}.
\newblock \href {https://doi.org/10.18653/v1/2023.emnlp-main.470} {{W}i{CE}: Real-world entailment for claims in {W}ikipedia}.
\newblock In \emph{Proceedings of the 2023 Conference on Empirical Methods in Natural Language Processing}, pages 7561--7583, Singapore. Association for Computational Linguistics.

\bibitem[{Kamoi et~al.(2023{\natexlab{b}})Kamoi, Goyal, Rodriguez, and Durrett}]{kamoi2023wicerealworldentailmentclaims}
Ryo Kamoi, Tanya Goyal, Juan~Diego Rodriguez, and Greg Durrett. 2023{\natexlab{b}}.
\newblock \href {https://arxiv.org/abs/2303.01432} {Wice: Real-world entailment for claims in wikipedia}.
\newblock \emph{Preprint}, arXiv:2303.01432.

\bibitem[{Kim et~al.(2023)Kim, Kim, Jeon, Park, and Kang}]{kim2023treeclarificationsansweringambiguous}
Gangwoo Kim, Sungdong Kim, Byeongguk Jeon, Joonsuk Park, and Jaewoo Kang. 2023.
\newblock \href {https://arxiv.org/abs/2310.14696} {Tree of clarifications: Answering ambiguous questions with retrieval-augmented large language models}.
\newblock \emph{Preprint}, arXiv:2310.14696.

\bibitem[{Klie et~al.(2024)Klie, de~Castilho, and Gurevych}]{10.1162/coli_a_00516}
Jan-Christoph Klie, Richard~Eckart de~Castilho, and Iryna Gurevych. 2024.
\newblock \href {https://doi.org/10.1162/coli_a_00516} {{Analyzing Dataset Annotation Quality Management in the Wild}}.
\newblock \emph{Computational Linguistics}, pages 1--50.

\bibitem[{Kotonya and Toni(2020)}]{kotonya2020explainableautomatedfactcheckingpublic}
Neema Kotonya and Francesca Toni. 2020.
\newblock \href {https://arxiv.org/abs/2010.09926} {Explainable automated fact-checking for public health claims}.
\newblock \emph{Preprint}, arXiv:2010.09926.

\bibitem[{Landis and Koch(1977)}]{Landis1977TheMO}
J~Richard Landis and Gary~G. Koch. 1977.
\newblock \href {https://api.semanticscholar.org/CorpusID:11077516} {The measurement of observer agreement for categorical data.}
\newblock \emph{Biometrics}, 33 1:159--74.

\bibitem[{Marshall et~al.(2020)Marshall, Nye, Kuiper, Noel-Storr, Marshall, Maclean, Soboczenski, Nenkova, Thomas, and Wallace}]{10.1093/jamia/ocaa163}
Iain~J Marshall, Benjamin Nye, Joël Kuiper, Anna Noel-Storr, Rachel Marshall, Rory Maclean, Frank Soboczenski, Ani Nenkova, James Thomas, and Byron~C Wallace. 2020.
\newblock \href {https://doi.org/10.1093/jamia/ocaa163} {Trialstreamer: A living, automatically updated database of clinical trial reports}.
\newblock \emph{Journal of the American Medical Informatics Association}, 27(12):1903--1912.

\bibitem[{Marshall et~al.(2021)Marshall, L'Esperance, Marshall, Thomas, Noel-Storr, Soboczenski, Nye, Nenkova, and Wallace}]{marshall2021state}
Iain~James Marshall, Veline L'Esperance, Rachel Marshall, James Thomas, Anna Noel-Storr, Frank Soboczenski, Benjamin Nye, Ani Nenkova, and Byron~C Wallace. 2021.
\newblock State of the evidence: a survey of global disparities in clinical trials.
\newblock \emph{BMJ Global Health}, 6(1):e004145.
\newblock PMCID: PMC7786802.

\bibitem[{Micallef et~al.(2022)Micallef, Armacost, Memon, and Patil}]{micallef2022true}
Nicholas Micallef, Vivienne Armacost, Nasir Memon, and Sameer Patil. 2022.
\newblock True or false: Studying the work practices of professional fact-checkers.
\newblock \emph{Proceedings of the ACM on Human-Computer Interaction}, 6(CSCW1):1--44.

\bibitem[{Min et~al.(2023)Min, Krishna, Lyu, Lewis, Yih, Koh, Iyyer, Zettlemoyer, and Hajishirzi}]{min-etal-2023-factscore}
Sewon Min, Kalpesh Krishna, Xinxi Lyu, Mike Lewis, Wen-tau Yih, Pang Koh, Mohit Iyyer, Luke Zettlemoyer, and Hannaneh Hajishirzi. 2023.
\newblock \href {https://doi.org/10.18653/v1/2023.emnlp-main.741} {{FA}ct{S}core: Fine-grained atomic evaluation of factual precision in long form text generation}.
\newblock In \emph{Proceedings of the 2023 Conference on Empirical Methods in Natural Language Processing}, pages 12076--12100, Singapore. Association for Computational Linguistics.

\bibitem[{Moberg et~al.(2018)Moberg, Oxman, Rosenbaum, Sch{\"u}nemann, Guyatt, Flottorp, Glenton, Lewin, Morelli, Rada, Alonso-Coello, and {GRADE Working Group}}]{Moberg2018-dx}
Jenny Moberg, Andrew~D Oxman, Sarah Rosenbaum, Holger~J Sch{\"u}nemann, Gordon Guyatt, Signe Flottorp, Claire Glenton, Simon Lewin, Angela Morelli, Gabriel Rada, Pablo Alonso-Coello, and {GRADE Working Group}. 2018.
\newblock The {GRADE} evidence to decision ({EtD}) framework for health system and public health decisions.
\newblock \emph{Health Res. Policy Syst.}, 16(1):45.

\bibitem[{Mohr et~al.(2022)Mohr, W{\"u}hrl, and Klinger}]{mohr-etal-2022-covert}
Isabelle Mohr, Amelie W{\"u}hrl, and Roman Klinger. 2022.
\newblock \href {https://aclanthology.org/2022.lrec-1.26} {{C}o{VERT}: A corpus of fact-checked biomedical {COVID}-19 tweets}.
\newblock In \emph{Proceedings of the Thirteenth Language Resources and Evaluation Conference}, pages 244--257, Marseille, France. European Language Resources Association.

\bibitem[{Nabo{\.{z}}ny et~al.(2021)Nabo{\.{z}}ny, Balcerzak, Wierzbicki, Morzy, and Chlabicz}]{info:doi/10.2196/26065}
Aleksandra Nabo{\.{z}}ny, Bart{\l}omiej Balcerzak, Adam Wierzbicki, Miko{\l}aj Morzy, and Ma{\l}gorzata Chlabicz. 2021.
\newblock \href {https://doi.org/10.2196/26065} {Active annotation in evaluating the credibility of web-based medical information: Guidelines for creating training data sets for machine learning}.
\newblock \emph{JMIR Med Inform}, 9(11):e26065.

\bibitem[{Pan et~al.(2023)Pan, Wu, Lu, Luu, Wang, Kan, and Nakov}]{pan-etal-2023-fact}
Liangming Pan, Xiaobao Wu, Xinyuan Lu, Anh~Tuan Luu, William~Yang Wang, Min-Yen Kan, and Preslav Nakov. 2023.
\newblock \href {https://doi.org/10.18653/v1/2023.acl-long.386} {Fact-checking complex claims with program-guided reasoning}.
\newblock In \emph{Proceedings of the 61st Annual Meeting of the Association for Computational Linguistics (Volume 1: Long Papers)}, pages 6981--7004, Toronto, Canada. Association for Computational Linguistics.

\bibitem[{Quelle and Bovet(2024)}]{10.3389/frai.2024.1341697}
Dorian Quelle and Alexandre Bovet. 2024.
\newblock \href {https://doi.org/10.3389/frai.2024.1341697} {The perils and promises of fact-checking with large language models}.
\newblock \emph{Frontiers in Artificial Intelligence}, 7.

\bibitem[{Ratcliff et~al.(2021)Ratcliff, Wong, Jensen, and Kaphingst}]{ratcliff2021impact}
Chelsea~L Ratcliff, Bob Wong, Jakob~D Jensen, and Kimberly~A Kaphingst. 2021.
\newblock The impact of communicating uncertainty on public responses to precision medicine research.
\newblock \emph{Annals of Behavioral Medicine}, 55(11):1048--1061.

\bibitem[{Saakyan et~al.(2021)Saakyan, Chakrabarty, and Muresan}]{saakyan-etal-2021-covid}
Arkadiy Saakyan, Tuhin Chakrabarty, and Smaranda Muresan. 2021.
\newblock \href {https://doi.org/10.18653/v1/2021.acl-long.165} {{COVID}-fact: Fact extraction and verification of real-world claims on {COVID}-19 pandemic}.
\newblock In \emph{Proceedings of the 59th Annual Meeting of the Association for Computational Linguistics and the 11th International Joint Conference on Natural Language Processing (Volume 1: Long Papers)}, pages 2116--2129, Online. Association for Computational Linguistics.

\bibitem[{Sackett et~al.(1996)Sackett, Rosenberg, Gray, Haynes, and Richardson}]{Sackett71}
David~L Sackett, William M~C Rosenberg, J~A~Muir Gray, R~Brian Haynes, and W~Scott Richardson. 1996.
\newblock \href {https://doi.org/10.1136/bmj.312.7023.71} {Evidence based medicine: what it is and what it isn{\textquoteright}t}.
\newblock \emph{BMJ}, 312(7023):71--72.

\bibitem[{Sarrouti et~al.(2021)Sarrouti, Ben~Abacha, Mrabet, and Demner-Fushman}]{sarrouti-etal-2021-evidence-based}
Mourad Sarrouti, Asma Ben~Abacha, Yassine Mrabet, and Dina Demner-Fushman. 2021.
\newblock \href {https://doi.org/10.18653/v1/2021.findings-emnlp.297} {Evidence-based fact-checking of health-related claims}.
\newblock In \emph{Findings of the Association for Computational Linguistics: EMNLP 2021}, pages 3499--3512, Punta Cana, Dominican Republic. Association for Computational Linguistics.

\bibitem[{Sim and Wright(2005)}]{10.1093/ptj/85.3.257}
Julius Sim and Chris~C Wright. 2005.
\newblock \href {https://doi.org/10.1093/ptj/85.3.257} {The kappa statistic in reliability studies: Use, interpretation, and sample size requirements}.
\newblock \emph{Physical Therapy}, 85(3):257--268.

\bibitem[{Simpkin and Armstrong(2019)}]{simpkin2019communicating}
Arabella~L Simpkin and Katrina~A Armstrong. 2019.
\newblock Communicating uncertainty: a narrative review and framework for future research.
\newblock \emph{Journal of general internal medicine}, 34:2586--2591.

\bibitem[{Thoma and Eaves(2015)}]{10.1093/asj/sjv130}
Achilleas Thoma and III Eaves, Felmont~F. 2015.
\newblock \href {https://doi.org/10.1093/asj/sjv130} {A brief history of evidence-based medicine (ebm) and the contributions of dr david sackett}.
\newblock \emph{Aesthetic Surgery Journal}, 35(8):NP261--NP263.

\bibitem[{Vladika et~al.(2024)Vladika, Schneider, and Matthes}]{vladika-etal-2024-healthfc}
Juraj Vladika, Phillip Schneider, and Florian Matthes. 2024.
\newblock \href {https://aclanthology.org/2024.lrec-main.709} {{H}ealth{FC}: Verifying health claims with evidence-based medical fact-checking}.
\newblock In \emph{Proceedings of the 2024 Joint International Conference on Computational Linguistics, Language Resources and Evaluation (LREC-COLING 2024)}, pages 8095--8107, Torino, Italia. ELRA and ICCL.

\bibitem[{Vykopal et~al.(2024)Vykopal, Pikuliak, Ostermann, and Šimko}]{vykopal2024generativelargelanguagemodels}
Ivan Vykopal, Matúš Pikuliak, Simon Ostermann, and Marián Šimko. 2024.
\newblock \href {https://arxiv.org/abs/2407.02351} {Generative large language models in automated fact-checking: A survey}.
\newblock \emph{Preprint}, arXiv:2407.02351.

\bibitem[{Wadhwa et~al.(2023)Wadhwa, Khetan, Amir, and Wallace}]{wadhwa-etal-2023-redhot}
Somin Wadhwa, Vivek Khetan, Silvio Amir, and Byron Wallace. 2023.
\newblock \href {https://doi.org/10.18653/v1/2023.findings-eacl.61} {{R}ed{HOT}: A corpus of annotated medical questions, experiences, and claims on social media}.
\newblock In \emph{Findings of the Association for Computational Linguistics: EACL 2023}, pages 809--827, Dubrovnik, Croatia. Association for Computational Linguistics.

\bibitem[{Wang et~al.(2024)Wang, Wu, Ke, Gao, Xu, and Chen}]{wang2024interactivemultimodalqueryanswering}
Mengzhao Wang, Haotian Wu, Xiangyu Ke, Yunjun Gao, Xiaoliang Xu, and Lu~Chen. 2024.
\newblock \href {https://arxiv.org/abs/2407.04217} {An interactive multi-modal query answering system with retrieval-augmented large language models}.
\newblock \emph{Preprint}, arXiv:2407.04217.

\bibitem[{Wanner et~al.(2024)Wanner, Ebner, Jiang, Dredze, and Van~Durme}]{wanner-etal-2024-closer}
Miriam Wanner, Seth Ebner, Zhengping Jiang, Mark Dredze, and Benjamin Van~Durme. 2024.
\newblock \href {https://doi.org/10.18653/v1/2024.starsem-1.13} {A closer look at claim decomposition}.
\newblock In \emph{Proceedings of the 13th Joint Conference on Lexical and Computational Semantics (*SEM 2024)}, pages 153--175, Mexico City, Mexico. Association for Computational Linguistics.

\bibitem[{Warren et~al.(2025)Warren, Shklovski, and Augenstein}]{10.1145/3706598.3713277}
Greta Warren, Irina Shklovski, and Isabelle Augenstein. 2025.
\newblock \href {https://doi.org/10.1145/3706598.3713277} {Show me the work: Fact-checkers' requirements for explainable automated fact-checking}.
\newblock In \emph{Proceedings of the 2025 CHI Conference on Human Factors in Computing Systems}, CHI '25, New York, NY, USA. Association for Computing Machinery.

\bibitem[{Zhai(2020)}]{zhai-2020-interactive}
ChengXiang Zhai. 2020.
\newblock \href {https://doi.org/10.1145/3397271.3401424} {Interactive information retrieval: Models, algorithms, and evaluation}.
\newblock In \emph{Proceedings of the 43rd International {ACM} {SIGIR} conference on research and development in Information Retrieval, {SIGIR} 2020, Virtual Event, China, July 25-30, 2020}, pages 2444--2447. {ACM}.

\bibitem[{Zhang and Choi(2023)}]{zhang2023clarifynecessaryresolvingambiguity}
Michael J.~Q. Zhang and Eunsol Choi. 2023.
\newblock \href {https://arxiv.org/abs/2311.09469} {Clarify when necessary: Resolving ambiguity through interaction with lms}.
\newblock \emph{Preprint}, arXiv:2311.09469.

\bibitem[{Zhang et~al.(2024{\natexlab{a}})Zhang, Knox, and Choi}]{zhang2024modelingfutureconversationturns}
Michael J.~Q. Zhang, W.~Bradley Knox, and Eunsol Choi. 2024{\natexlab{a}}.
\newblock \href {https://arxiv.org/abs/2410.13788} {Modeling future conversation turns to teach llms to ask clarifying questions}.
\newblock \emph{Preprint}, arXiv:2410.13788.

\bibitem[{Zhang et~al.(2024{\natexlab{b}})Zhang, Qin, Deng, Huang, Lei, Liu, Jin, Liang, and Chua}]{zhang2024clamberbenchmarkidentifyingclarifying}
Tong Zhang, Peixin Qin, Yang Deng, Chen Huang, Wenqiang Lei, Junhong Liu, Dingnan Jin, Hongru Liang, and Tat-Seng Chua. 2024{\natexlab{b}}.
\newblock \href {https://arxiv.org/abs/2405.12063} {Clamber: A benchmark of identifying and clarifying ambiguous information needs in large language models}.
\newblock \emph{Preprint}, arXiv:2405.12063.

\end{thebibliography}

\clearpage
\appendix

\onecolumn
\section{Trikafta Claim Example}
\label{sec:trikafta_example}
\begin{table*}[h]
\scriptsize
\begin{longtable}{>
{\raggedright\arraybackslash}p{3.2cm} >{\raggedright\arraybackslash}p{5.39cm} >{\raggedright\arraybackslash}p{2cm} >{\raggedright\arraybackslash}p{1.8cm} >{\raggedright\arraybackslash}p{2.4cm} }
    \toprule
        \multicolumn{3}{p{\dimexpr4cm+4cm+4cm}} 
        {\RHExample{Dandruff and Trikafta}{Anyone had really \hlgreen{bad flaky scalp} or \hlgreen{dandruff} lately ? \hlyellow{Think it could be due to }\hlpink{trikafta}\hlyellow{.} Could it be anything else}{Patients with Cystic Fibrosis (implied by the subreddit r/CysticFibrosis)}{Trikafta (a medication)}{Flaky scalp or dandruff}{r/CysticFibrosis}} 
        \vspace{0.3em} \\

    \midrule

\cellcolor{lightgray!20}\tiny\textbf{Expert 1:} None of the abstracts directly addressed patients with cystic fibrosis experiencing dandruff/skin-related adverse effects secondary to trikafta use. All of the abstracts include some form of scalp flaking whether it be dandruff in general or specific conditions such as seborrheic dermatitis. However, they cannot be considered relevant as none of them address patients with cystic fibrosis or experiencing dandruff secondary to medication side effect.
    &\cellcolor{lightgray!20}\tiny\textbf{Expert 2:} A1, a2, a3, a4, a6 include the treatment of dandruff in a population with dandruff. The only relevant element in these abstracts is the outcome measures.

 A5, a7 includes a population with psoriasis and therefore, even the outcome measure here is irrelevant (i.e., all PIOs irrelevant). Similarly, a8 and a 9 included patients with seborrhoeic dermatitis, whereby the population and intervention were irrelevant, but scaling was an outcome measure (I would suggest somewhat relevant outcome). 

A10 also involved psoriasis population and intervention, but outcomes included scaling, therefore partially relevant outcome.

Overall, the results are entirely inconclusive, since no abstract was relevant.
 &\cellcolor{lightgray!20}\tiny\textbf{Expert 3:} Overall, no relevant abstracts were available to analyze if trikafta may cause bad flaky scalp or dandruff. This is a very specific claim, and it is usually verified in the side effects results of RCTs. If not asked, participants may ignore the symptom if it is not significant. \\
\cellcolor{lightgray!20}\tiny\textbf{Expert 4:} There are no relevant abstracts to determine overall support. None of them include cystic fibrosis patients or the medication (Trikafta) from the claim. The outcome is mentioned in abstracts a1, a3, a4, a6, but has no relation to the claim.
&\cellcolor{lightgray!20}\tiny\textbf{Expert 5:} The available studies included individuals with dandruff however none were diagnosed with cystic fibrosis, none were given Trikafta (a1, a2, a3, a4, a5, a6, a7, a8, a9, a10).

Expert opinion: Dandruff can be caused by the underlying condition (cystic fibrosis) rather than as an effect of the medication itself (Trikafta)
\\
\bottomrule
\caption{All experts found the claim unverifiable based on the available RCTs. They attributed this to the claim’s high specificity, noting it is unlikely—and potentially unethical—for a trial to match the described scenario.}

\label{table:trikafta_example}\\
\end{longtable}
\end{table*}
\normalsize

\newpage
\section{All Claims from Final Split}
\begin{table*}[h]
    \centering
    \scriptsize
\begin{tabular}{
p{0.97\textwidth}} \toprule
        \RHExample{ADHD, Herbs, and Menstruation}{Hello, menstruating people! How do your cycle and ADHD influence each other and how do you deal with it?

EDIT: After getting your responses I am reflecting again how medicine does not give a shit about women. It's truly insane. Thank you!

Hello! I have never paid too much attention to my menstrual cycle since it was never particularly bothersome. Now that I take methylo I feel big changes in how I function during the cycle. Like last 10 days of the cycle, my medication kind of stops working... That is like 1/3 of the time. I know it's still better than without meds nevertheless, it makes establishing a routine quite challenging. My doc suggested trying contrac\hlpink{eptive} pills, but I am not even sexually \hlgreen{active ATM so taking} more medication, with potential side effects, does not excite me.

\hlyellow{I know there are herbs that are proven to be helping with regulating the cycle} but I don't know if they would help with ADHD symptoms? Any tips?}{People with ADHD}{Herbs}{Regulating the menstrual cycle}{r/ADHD}
        \vspace{0.3em} \\
        \rowcolor{gray!10}
        \RHExample{Stimulants and Sodium}{
Stimulants vs. Sodium

Im wondering if anyone else has experienced this. \hl{I find that my stimulant medications (}\hlpink{Adderall IR}\hlyellow{ and }\hlpink{Vyvanse}\hlyellow{) make me very sensitive to }\hlgreen{salt}. If I have a higher sodium meal (eg ramen or canned soup, or even just mustard on my sandwich), I get very bloated. Its uncomfortable and lasts for a few days. Whenever I take a break from my meds, this doesnt happen. Ive had labs done for it in the past and it doesnt seem like anything medically problematic, but its uncomfortable and it really stresses me out.}{People with ADHD}{Stimulant medications (Adderall IR and Vyvanse)}{Sensitivity to salt (resulting in bloating)}{r/ADHD}

\vspace{0.3em} \\
        \RHExample{Dandruff and Trikafta}{Anyone had really \hlgreen{bad flaky scalp} or \hlgreen{dandruff} lately ? \hlyellow{Think it could be due to }\hlpink{trikafta}\hlyellow{.} Could it be anything else}{Patients with Cystic Fibrosis (implied by the subreddit r/CysticFibrosis)}{Trikafta (a medication)}{Flaky scalp or dandruff}{r/CysticFibrosis}

        \vspace{0.3em} \\
        \rowcolor{gray!10}
        \RHExample{Pineapple Juice Reduces Inflammation}{Anyone with sinus issues drinking pineapple juice?

It's a weird question, but \hl{I saw a post about }\hlpink{pineapple}\hlyellow{ juice being good for sinus issues (helps with the }\hlblue{inflammation}\hlyellow{)} and just wondered if anyone has done this? Some people were commenting about the high sugar content in pineapple juice not being good, but they get around that by taking a supplement instead of drinking the juice. Anyone?
}{Patients With Cystic Fibrosis}{Pineapple Juice}{Reduced Inflammation/Fewer Sinus Issues}{r/CysticFibrosis}

\vspace{0.3em} \\    

        \RHExample{Trikafta and PMDD}{Trikafta \& PMDD

So, \hl{I believe }\hlpink{trikafta}\hlyellow{ has given me }\hlpink{PMDD}\hlyellow{ premenstrual dysphagia disorder}. Every month, the week before my period I have extreme \hlgreen{anxiety} in a running dialouge in my head that is constantly negative. I've never been this way before. I also have horrible \hlpink{hormonal acne} on my back \& forehead which are very new to me as well.

My question is: any one else having this problem? My Dr said they are noticing a "negative interaction with estrogen and trikafta". Anyone find anything that helps??}{Patients with cystic fibrosis (implied by the Reddit thread r/CysticFibrosis), specifically females of reproductive age}{Trikafta}{
\begin{varwidth}{0.75\textwidth}
Development of PMDD (premenstrual dysphoric disorder, not dysphagia disorder) symptoms, including extreme anxiety and hormonal acne.
\end{varwidth}
}{r/CysticFibrosis} 
        \vspace{0.3em} \\

    \bottomrule
\end{tabular}
\vspace{-0.5em}
\caption{All claims from the final annotation split. The extracted claim span is highlighted in \hl{yellow}; Population annotations are highlighted in \hlblue{blue}, Intervention in \hlpink{pink}, and Outcome in \hlgreen{green}.
}
\label{table:10_18_full}
\end{table*}

\newpage
\twocolumn
\section{Rationale for Re-extracting PIO Elements}
\label{appendix:pioelements}

During the pilot and refinement rounds, we identified a key source of expert disagreement arising from underspecified focus when multiple Population, Intervention, and Outcome (PIO) elements were present in a claim. In such cases, experts differed on whether to consider all PIO elements or prioritize a subset, leading to inconsistent judgments. An illustrative example is shown in Table~\ref{table:prednisone_ex}.

Although RedHOT provides PIO highlights for each claim, we observed occasional errors and inconsistencies in these annotations (e.g., mislabeling a Population as an Intervention), which could further contribute to ambiguity during expert review. To reduce this source of disagreement and standardize expert attention, we re-extracted explicit PIO elements from each post using an LLM-based pipeline, with prompt design and expert validation detailed in Appendix~\ref{appendix:pioextraction}.

\section{Discussion on the Evaluation of the Communication Model}

In proposing the communication model, we also contemplated over how such a system would be evaluated.
We considered patient-system alignment and expert-system alignment to be of particular importance. The system needs to be aligned on the patient’s information needs and on the correct information, as would be given by a medical expert, to provide to fulfill those needs. Evaluating along these axes would likely necessitate a benchmark, created in cooperation with medical experts, in which a system engages in conversations with a simulated patient that is conditioned to have certain information needs. The system would then be evaluated on whether it can identify and correctly address those needs. The details for such an evaluation would be up to future work and is out of the scope for this paper.

\scriptsize
\begin{table*}[h]
\begin{tabular}{p{0.3\textwidth} p{0.3\textwidth} p{0.3\textwidth}}

    \toprule
        \multicolumn{3}{p{0.9\textwidth}}        
        {\scriptsize
        \ClaimExample{Prednisone}{Cytoxan and prednisone

Rheumatologist says \hlpink{cellcept} failed to protect my kidneys and now I have developed \hlgreen{lupus nephritis}.Im so upset. \hlpink{Prednisone} messed up my hips so badly that they both need to be replacedI dont want to get back on it but \hlyellow{rheumatologist says its to bring the }\hlgreen{inflammation}\hlyellow{ down in my kidneys.} Ive never been on \hlpink{Cytoxan} but the side effects sound identical to a lupus flare. How am I supposed to be positive with news like this? I feel so defeatedI dont know what to do.}{r/lupus}}  \\      
     \midrule

\cellcolor{green!20}\tiny\textbf{Expert 1:} Overall consensus of relevant abstracts is that the combination of prednisone with cyclophosphamide (Cytoxan) is effective in treating kidney inflammation due to lupus nephritis (a3, a6, a8, a9, 10). This is in support of the original claim. However, one abstract found that kidney function continues to gradually deteriorate even with treatment (a2). Due to the majority of abstracts supporting the original claim however, the conclusion can be made that cyclophosphamide and prednisoine combination therapy is effective for decreasing renal inflammation in lupus nephritis.

&\cellcolor{orange!20}\tiny\textbf{Expert 2:} The overall conclusion is that the claim is partially refuted. Prednisone alone appears to lead to renal deterioration (a2, a3, a6, a7, a8), but in combination with immunosuppressants, can have beneficial effects (a9). 
Abstracts a4 and a5 were somewhat relevant (did not specifically test prednisone effects).
Irrelevant abstracts included a1 and a10.
All relevant abstracts included lupus nephritis as the population. The person who made the claim appears to have arthritis plus lupus nephritis, therefore, none of the abstracts reported this exact population.

 &\cellcolor{lightgray!20}\tiny\textbf{Expert 3:} None of the given abstracts were relevant to verify the claim.
 
None of the studies had a control group for prednison treatment in lupus nephritis. All groups in all of the given studies recieved prednison as a base treatment and compared this to the effects of an additional immunosuppressive drug. 
Since lupus (-nephritis) is an autoinflammatory disease, it is usually (depending on the severity) treated with immunosuppressive glucocorticoids such as prednison to inhibit the autodestruction of tissues and organs.\\

\cellcolor{lime!20}\tiny\textbf{Expert 4:} Overall, the abstracts partially support the use of prednisone to reduce kidney inflammation. With the exception of study a7, every other study included either prednisone or glucocorticoid in both the treatment and control groups. The difference usually is between glucocorticoid only or low-dose. Even a7, the only one that does not show a low dose of glucocorticoid use, might not appear to do so because methods may not be totally revealed in the abstract. Therefore, the abstracts suggest that this patient might receive a prescription for at least low-dose prednisone.

&\cellcolor{green!20}\tiny\textbf{Expert 5:} Core: The evidence supports the benefit of immunosuppressive medications such as Cyclophosphamide in addition to steroids and oral maintenance medications for those with lupus nephritis. 

Addendum: There is no study to support the superiority of Cyclophosphamide over other immunosuppressive medications especially if the patient has already had a poor response to other immunosuppressive medications such as Mycophenolate mofetil.

&\cellcolor{lime!20}\tiny\textbf{Expert 6:} Abstracts a9, a3 a10, a6, a8 conclude that de combination of cytoxan and glucocorticoids shows better outcomes in patients with nephritis related to systemic lupus, which are two of the meds mentioned in the claim. 

The abstracts a7, a5, a2 are somewhat relevant.

Since the majority of the abstracts did mentioned better outcomes using the medications from the claim, but cytoxan wasn’t the only immunosuppressive drug compared in the studies, I’d say overall the abstracts partially support.\\
\bottomrule
\end{tabular}
\caption{Example illustrating expert disagreement when multiple PIO elements are present and no explicit guidance is provided on which elements to prioritize.}\label{table:prednisone_ex}
\end{table*}

\normalsize

\twocolumn
\section{Changes between annotation rounds}
\label{appendix:roundschanges}
In the first round, ten claims were annotated without explicit PIO contextualization or the expert support field. In the second round, we refined the annotation guidelines to clarify label definitions; however, agreement on the refinement set of five claims remained low.
Based on expert feedback, we substantially revised the setup for the third round by filtering out non–RCT-verifiable claims, re-extracting and providing PIO elements to guide expert focus, improving the retrieval system, and further clarifying the annotation guidelines. Experts then annotated five claims under this updated setting.

\section{Annotation Guidelines}
\label{appendix:annoguidelines}
\normalsize

We present the guideline given to our expert annotators below. These instructions were given in slide-deck format to annotators with images from the annotation interface spliced in-between to clearly indicate how to annotate.

\tiny
\textbf{The Task}\\
This annotation task involves verifying medical claims made on Reddit posts using retrieved evidence. You will be looking at the provided abstracts to determine whether, when considering all the evidence, you can support or refute the claim.\\

\textbf{Post}\\
We will give you a Reddit post, which is annotated to contain the following.
\begin{itemize}
    \item What subreddit the post is from.
    \item Spans indicating PIO (Population, Intervention, Outcome) elements.
    \begin{itemize}
        \item Population indicates the affected subjects (ex: COVID patients, diabetics).
        \item Intervention indicates any treatments applied to the subjects (ex: remdesivir, Ozempic).
        \item Outcome indicates how the effects of the intervention are evaluated (ex: pain, weight, 30 day mortality).
    \end{itemize}
    \item Claim Span: Part of the post that makes the medical claim that we analyze.
\end{itemize}

\textbf{Post \& Derived Claim}
\begin{itemize}
    \item You will NOT be directly evaluating the information in the post. It is presented to you as to inform you of the context in which the claim is made.
    \item What you will be directly evaluating is the claim derived from the post. We present you with a (P, I, O) tuple extracted from the post that we use to make the claim as clear as possible.
    \item In some cases, the claim in the post may be ambiguous. In this case, we will present a disambiguated claim for you to evaluate.\\
\end{itemize}

\textbf{RCT-Verifiability}
\begin{itemize}
    \item A claim is RCT-Verifiable if there exists (or should exist) a reasonable RCT that will be able to either support or refute it.
        \begin{itemize}
            \item A reasonable RCT is one that can be practically and ethically conducted.
        \end{itemize}
    \item Most of the claims we give should be RCT-Verifiable. However, this may not always be the case.
    \item In the case when a claim is not RCT-Verifiable, you should indicate as such.
        \begin{itemize}
            \item You will be forced to write a 10 word explanation for why the claim is not RCT-Verifiable. Please ensure that the explanation is for a legitimate reason, for example, an unethical intervention, as we will review. You can only continue to annotate once you are done.
        \end{itemize}
\end{itemize}

\textbf{Retrieved Abstracts}
\begin{itemize}
    \item For each claim, you are given 10 abstracts that are retrieved automatically based on information in the claim.
    \item Each abstract has the following:
    \begin{itemize}
        \item Title
        \item Published Date
        \item Informative Highlights: PIO Spans and Abstract Punchline (Span describing the core of the abstract’s findings)
    \end{itemize}
    \item We provide you a way to flag abstracts that you believe to be of poor quality in the interface. Be sure to keep in mind the quality of the RCT experiment described in the abstract when annotating them.
\end{itemize}

\textbf{Relevance Annotations}
\begin{itemize}
    \item For these annotations, you will be analyzing the relevance of the abstract to the claim being evaluated.
    \item You will be analyzing the relevance of the following four components:
    \begin{itemize}
        \item Population: Is the population being studied in the abstract relevant to the population the claim is addressing?
        \item Intervention: Is the intervention being studied in the abstract relevant to the population the claim is addressing?
        \item Outcome: Are any of the outcome measures used in the RCT described in the abstract relevant to the population the claim is addressing?
        \item Overall: Is the abstract relevant enough to the claim for it to be used to verify the claim?
    \end{itemize}
    \item As mentioned before, you may flag the abstract if you think it is of concerning quality.
\end{itemize}

\textbf{Relevance Labels}
\begin{itemize}
    \item For each relevance component, you are given 4 labels to choose from. They are as follows:
    \begin{itemize}
        \item Select (Default)
        \begin{itemize}
            \item For PIO: The element is missing.
            \item For Overall: The abstract does not describe an RCT.
        \end{itemize}
    \end{itemize}
    \begin{itemize}
    \item Irrelevant
    \begin{itemize}
        \item For PIO: The element in the abstract has no relation at all to the corresponding element in the post.
        \begin{itemize}
            \item Ex for Population - Claim: Patients with LPR - Abstract: Healthy Patients
        \end{itemize}
        \item For Overall: No part of this abstract can be used to make even an inference on whether the claim can be supported or refuted.
    \end{itemize}    
    \end{itemize}
    
    \item Somewhat Relevant
    \begin{itemize}
        \item For PIO: Indicates that the element has some relation to the corresponding element in the post, but is not close enough for it to be used to directly verify the claim even if all other elements are 100\% relevant.
        \begin{itemize}
             \item Ex for Population - Claim: Patients with LPR - Abstract: Patients with gastro-oesophageal reflux disease
        \end{itemize}
        \item For Overall: Some parts of this abstract can be used to make an inference on whether the claim can be supported or refuted. However, this abstract still cannot be used as direct evidence to support or refute the claim.
    \end{itemize}
    \item Relevant
    \begin{itemize}
        \item For PIO: Indicates that the element in the abstract is close enough to the corresponding element in the post for the purpose of verifying the claim.
        \begin{itemize}
            \item Ex for Population - Claim: Patients with LPR - Abstract: Patients with laryngopharyngeal reflux
        \end{itemize}
        \item For Overall: The abstract can be used as direct evidence to support or refute the claim.
    \end{itemize}
\end{itemize}

\textbf{Abstract Support}

\begin{itemize}
    \item If overall, the abstract is relevant to the claim. You will be given the opportunity to annotate for whether the abstract supports/refutes the claim.
    \item There are four labels you can choose from:
    \begin{itemize}
        \item Refutes: This abstract fully refutes the claim in the post.
    \item Partially Refutes: This abstract refutes the claim given some condition or caveat.
    \item Partially Supports: This abstract supports the claim given some condition or caveat.
    \item Supports: This abstract supports the claim in the post.
    \end{itemize}
    \item A partial support or refute indicates that there is some nuance in the RCT result that prevents the abstract from fully supporting or refuting the claim.
    \begin{itemize}
        \item A sub-group of the population experienced different results from the rest.
    \item The results can only be reproduced under specific conditions that cannot be generalized.
    \end{itemize}
\end{itemize}

\textbf{Relevant Span}\\
When you are done determining the support label for abstract. You must determine which span of text in the abstract is most relevant in indicating whether the abstract can be supported or refuted.\\

\textbf{Tiering}
\begin{itemize}
    \item After you are done with annotating all the abstracts you can start the tiering and synthesis.
    \item In the tiering phase, you are organizing the abstracts into tiers.
    \begin{itemize}
        \item Abstracts are automatically tiered according to the relevance annotations.
        \item You should attempt to further categorize the abstract according to their quality or their importance regarding the claim, as well as temporal relevance (up your own medical expertise).
    \end{itemize}
\end{itemize}

\textbf{Synthesis}
\begin{itemize}
    \item For this task, you must pick the label determining whether the claim is supported or refuted according to two criteria.
    \begin{itemize}
        \item Overall Support (OS): Determine whether the claim is supported or refuted using only the provided evidence.
        \item Expert Opinion (EO): Determine whether the claim is supported or refuted using your expert knowledge.
    \end{itemize}
    \item You are given 6 labels to choose from:
    \begin{itemize}
        \item No Relevant Abstracts/No Expert Opinion:
        \begin{itemize}
             \item OS: There are no relevant abstracts to determine overall support.
            \item EO: You don’t have the expert knowledge in the field to make this decision.
        \end{itemize}
    \item Refutes:
    \begin{itemize}
        \item OS: Overall, considering all the abstracts, there is strong evidence that the claim can be refuted.
        \item EO: According to your expert knowledge, this claim can be strongly refuted
    \end{itemize}.
    \item Partially Refutes:
    \begin{itemize}
        \item OS: Overall, considering all the abstracts, there is evidence that the claim can be refuted depending on some general condition or caveat.
        \item EO: According to your expert knowledge, this claim can be refuted depending on some general condition or caveat.
    \end{itemize}
    \item Inconclusive:
    \begin{itemize}
        \item OS: This should rarely happen. Only pick this in cases, where there is true deadlock within the evidence as to whether the claim can be supported or refuted.
    \item EO: According to your expert knowledge, there is no scientific consensus that points to the claim being supported or refuted.
    \end{itemize}
    \item Partially Supports:
    \begin{itemize}
        \item OS: Overall, considering all the abstracts, there is evidence that the claim can be supported depending on some general condition or caveat.
    \item EO: According to your expert knowledge, this claim can be supported depending on some general condition or caveat.
    \end{itemize}
    \item Supports:
    \begin{itemize}
        \item OS: Overall, considering all the abstracts, there is strong evidence that the claim can be supported.
    \item EO: According to your expert knowledge, this claim can be strongly supported.
    \end{itemize}
    \end{itemize}
\end{itemize}

\textbf{Synthesis Explanation}

\begin{itemize}
    \item Afterwards, write an explanation of why you picked that option. Use the tiers you created earlier to help develop this explanation (quality, temporal, relevance). Cite the abstracts in your explanation.
    \begin{itemize}
        \item For example: There were a few abstracts that refuted the claim (a2, a5). …
    \end{itemize}
    \item Give sufficient context and details where someone can follow the reasoning without looking at your annotations. Think of the perspective of you explaining to a patient.
    \item Try and keep the grade level at around Middle School, try to avoid complex jargon
    \item We would like for you to include the following in your explanation:
    \begin{itemize}
        \item Main statement explaining why you selected the label.
    \item Rundown of how relevant evidence (abstracts) supports/conditionally supports/refutes claim.
    \item (optional) Addendum with relevant clinical experience regarding claim
    \item Use the terminology of a(abstract number) like a9 to refer to abstract 9 in your explanations.
    \item Aim for the length (without addendum) to be around ~100 words. If you don’t need that much explanation, 50 words is fine. If you really need to explain something in more detail, please keep it under ~150 words.
    \end{itemize}
    \item Please do not include:
    \begin{itemize}
        \item Any direct references to the tiers (Ex: The abstracts in tiers 1).
    \end{itemize}
\end{itemize}

\normalsize

\section{Plain Language Explanation Guideline}
\label{appendix:plainlanguageexplanationformat}
\begin{itemize}
\item Include an overall sentence either at the beginning or end of your synthesis explanation. 
\item Target to aim the explanations at 100 words or less, 150 words if there are details that must be elaborated on.
\item Include details of abstracts identified as relevant and explanations of how it supports the ultimate label, including some nuance.
\item (Optional) Medical Addendum at end.
\end{itemize}

\onecolumn
\section{Annotation Interface}
\label{appendix:annointerface}

\begin{figure}[h]
    \centering
    \includegraphics[width=\linewidth]{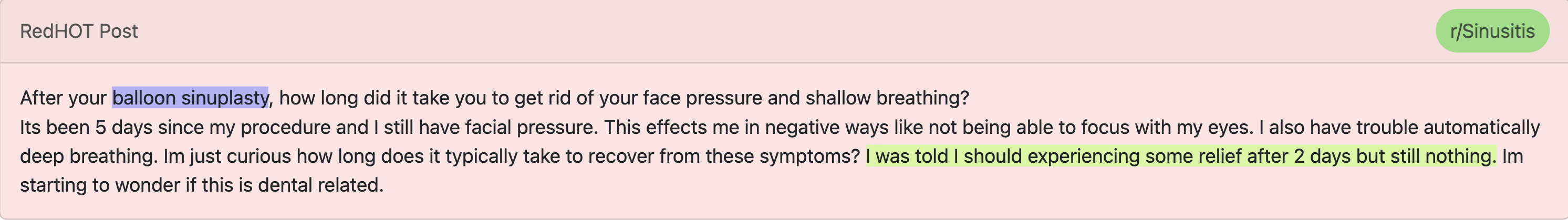}
    \includegraphics[width=\linewidth]{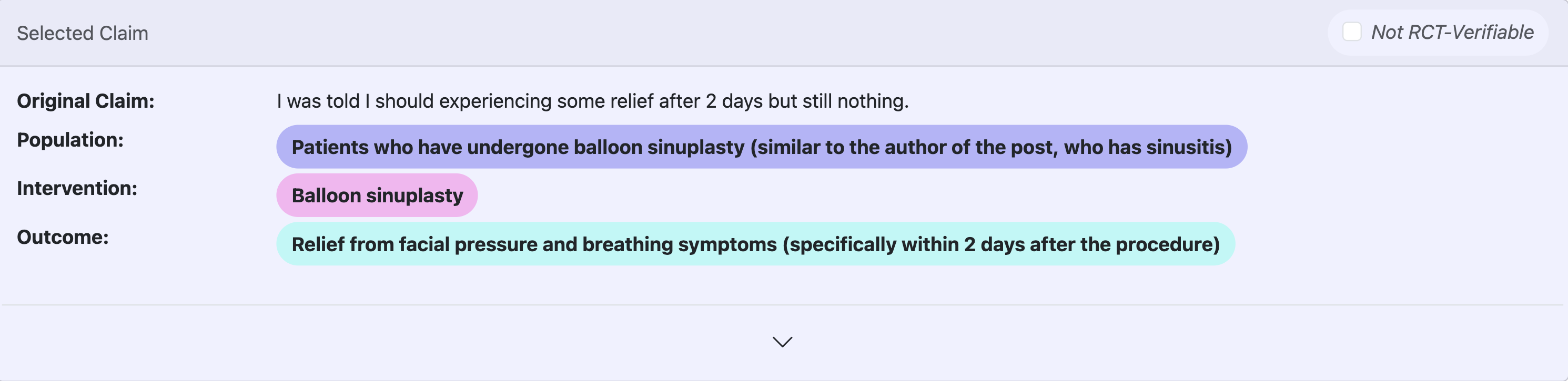}
    \includegraphics[width=\linewidth]{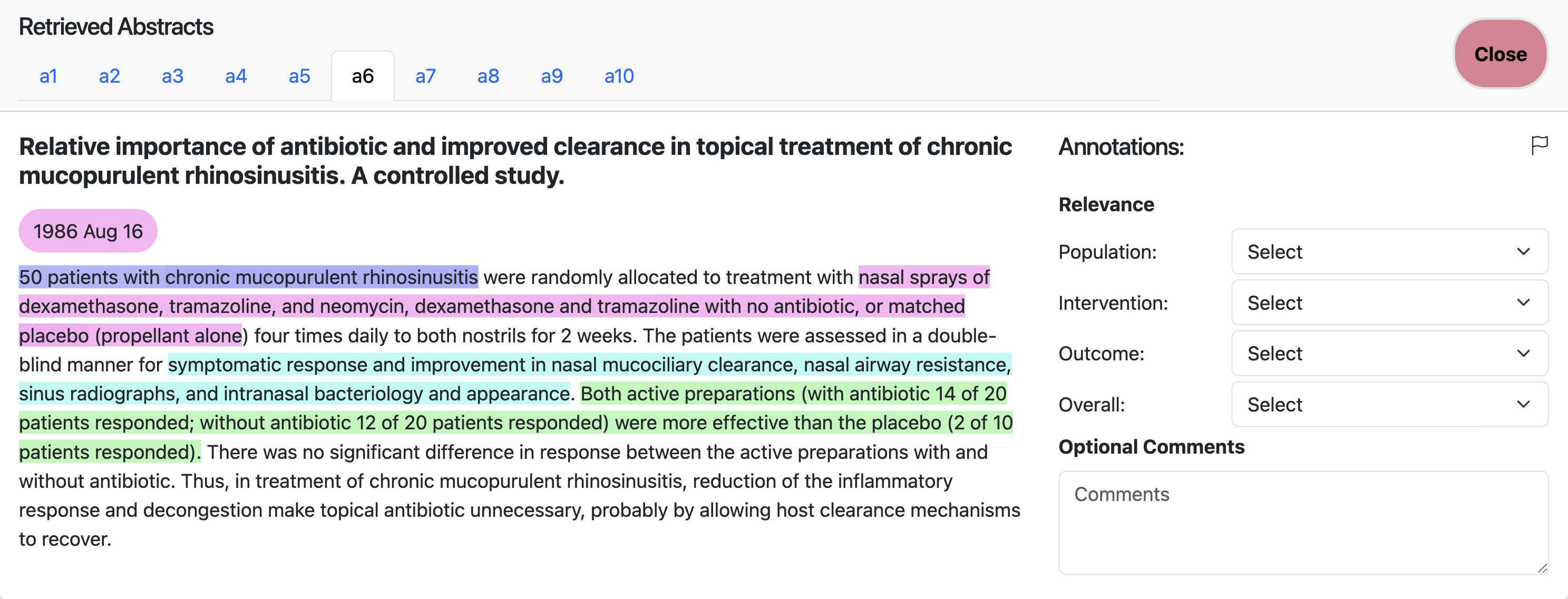}
    \caption{Presentation of claims, PIO elements, and abstracts in the annotation interface.}
    \label{fig:typical_view}
\end{figure}

\begin{figure}[h]
    \centering
    \includegraphics[width=\linewidth]{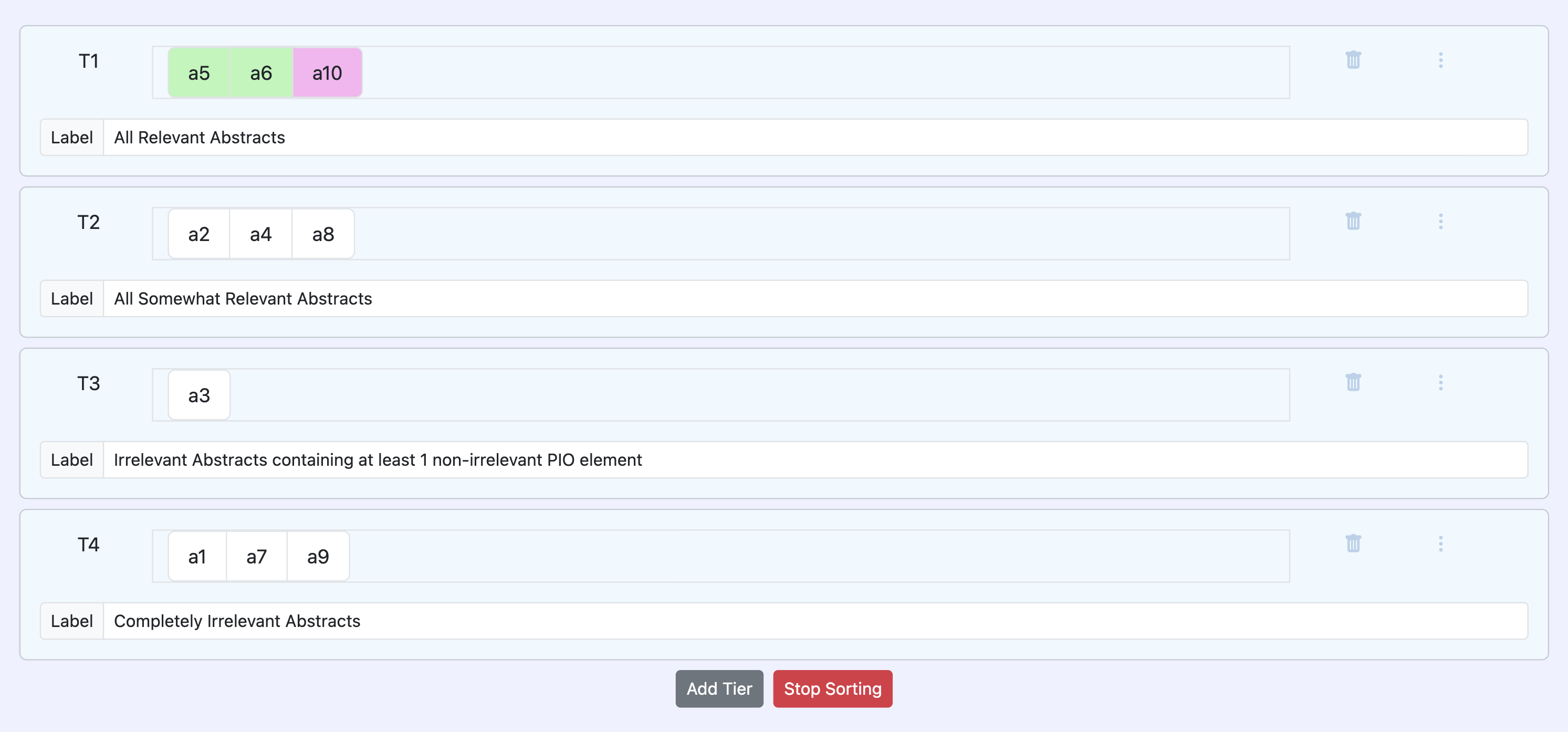}
    \includegraphics[width=\linewidth]{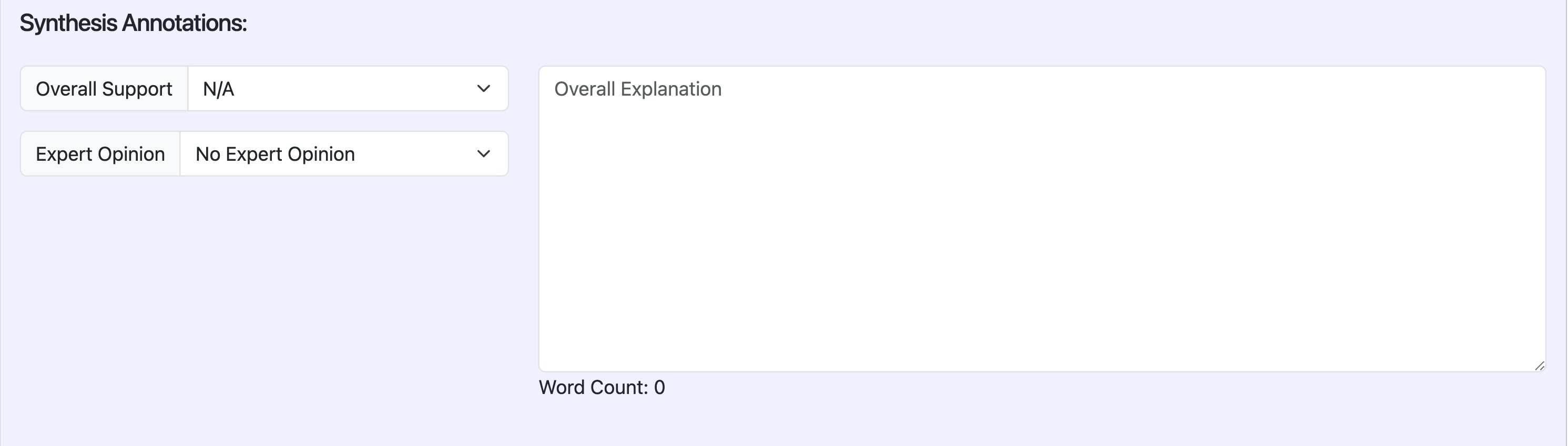}
    \caption{Presentation of the tiering and synthesis annotations interface.}
    \label{fig:synthesis_view}
\end{figure}

\twocolumn
We used a web-based annotation interface to collect annotations from our expert medical annotators. Figure~\ref{fig:typical_view} shows how we present a claim with its surrounding context, extracted PIO elements for that claim, and retrieved abstracts corresponding to that claim.
In this setup, we highlight the claim within the post in which it is found along with any PIO spans as determined by data found in the RedHOT dataset.
We also present extracted PIO elements (see Appendix~\ref{appendix:pioextraction}) in a separate box with a rewritten claim created by inputting these elements in a template. 
All of this information was provided for the benefit of the annotator to clearly understand the claim in question.

Reading each abstract can be a cumbersome task. Therefore, we also provide for each abstract, information highlights and the published date of the paper associated with that abstract.
These informational highlights covered the PIO elements in the abstract as well as the punchline of the abstract. This information is provided along with the abstracts in the TrialStreamer dataset. 

Figure~\ref{fig:synthesis_view} shows the interface in which the expert annotators would tier abstracts and then provide the overall synthesis annotations with explanations. After the annotator is done with their relevance annotations, the interface will automatically tier abstracts according to these annotations. These automatic tiers are:
\begin{itemize}
    \item \textbf{All Relevant Abstracts:} This tier contains all abstracts that were determined to be overall relevant to the claim.
    \item \textbf{All Somewhat Relevant Abstracts:} This tier contains all abstracts that were determined to be somewhat relevant to the claim.
    \item \textbf{Irrelevant Abstracts containing at least 1 non-irrelevant PIO element:} As the name suggests, this tier contains all irrelevant abstracts with at least 1 non-irrelevant PIO element in relation to the claim.
    \item \textbf{Completely Irrelevant Abstracts:} This tier contains all abstracts with the overall and all the PIO elements labeled as irrelevant in relation to the claim.
\end{itemize}

Expert annotators, when presented with these tiers, should try to further categorize the collection of abstracts if possible. They are able to add tiers, manipulate their order, and change their names. They can also double click on an abstract tag, and the interface will display the abstract corresponding to that tag. 
All of these features serve to make the process of tiering abstracts as streamlined as possible for the expert annotators. 

\section{Annotator Recruitment}
\label{appendix:annorecruitment}
We recruited five medical experts via Upwork over a four-week period. We received 117 proposals, screened 19 candidates using a sample annotation task (Table~\ref{table:gaviscon}) to assess instruction-following, medical knowledge, and explanation quality, and shortlisted seven candidates for interviews, from which five were selected.
Experts worked between 3 and 20 hours per week and required approximately 20 minutes to annotate each claim end-to-end. Compensation ranged from \$22 to \$35 per hour. The total cost of the annotation study was \$1{,}432.20.

\begin{table*}[h]
    \centering
    \scriptsize
\begin{tabular}{
p{0.97\textwidth}} \toprule
    \ClaimExample{Gaviscon Advance}{Can I buy liquid alginate suspension (Gaviscon Advance) in the U.S.?
Hi everyone. I'm newly diagnosed with LPR and doing a lot of research on the best treatments. \hl{I've read that a }\hlpink{liquid alginate suspension (Gaviscon Advance)}\hl{ is quite effective at treating }\hlblue{LPR} but it looks like it's not sold in the U.S. Does anyone know how I can find it here?}{r/GERD}
        \vspace{0.3em} \\
    \bottomrule
\end{tabular}
\vspace{-0.5em}
\caption{
}
\label{table:gaviscon}
\end{table*}

\section{PIO Extraction}
\label{appendix:pioextraction}

We used an automatic PIO extraction approach based on \verb|Llama-3.1-405B-Instruct|. After preliminary prompt testing and consultation with a medical expert, we incorporated PIO element definitions and guiding questions adapted from Duke University’s PICO evidence-based medicine guidelines\footnote{\url{https://guides.mclibrary.duke.edu/ebm/pico}}.

\medskip
\itshape
PATIENT OR PROBLEM\\
How would you describe a group of patients similar to yours? What are the most important characteristics of the patient?
Example: COVID patients, diabetics

INTERVENTION, EXPOSURE, PROGNOSTIC FACTOR\\
What main intervention, exposure, or prognostic factor are you considering? What do you want to do with this patient?
Example: Remdesivir, Ozempic

OUTCOME\\
What are you trying to accomplish, measure, improve or affect?
Example: pain, weight, 30 day mortality

Extract the Population, Intervention, and Outcome elements from the following claim from the following text.
Write “None” if the element does not exist in the text.

Text posted by someone in Reddit thread r/[sub][sub\_description]:

[post]

Highlighted claim:

[claim]
\normalfont
\medskip

On a validation set of 55 samples (five claims randomly sampled from each of 11 conditions), a medical expert reviewed the extracted PIO elements. The expert reported over 90\% accuracy for each element, noting only minor issues such as overlapping Population and Intervention spans and occasionally implied elements. Based on this review, the extraction pipeline was deemed sufficiently reliable for use in our study.

\section{Prompt for RCT Verifiability}
\label{appendix:rctpipeline}
\itshape
You are a potential clinical trialist. I will give you a claim and post. The claim is part of the post, and the post can give you context. I want you to tell me if the claim can be studied in a randomized controlled trial (RCT). An RCT can test an intervention to measure a benefit or non-inferiority. However, the RCT must be ethical: Ethical Guidelines: The intervention should not cause harm or have a significant risk of toxicity. It should not test exposures known to be potentially harmful, such as food-drug interactions that might cause adverse effects. The safety of participants is the primary concern, and interventions that pose significant health risks should not be tested in an RCT. Design Requirements: The trial needs to have a control group, with the only difference being the intervention. There must be a feasible and ethical way to measure outcomes without exposing participants to undue risk. Wait for my text to classify whether the claim can be ethically studied in an RCT.\\

Claim: [claim]\\

Text from r/[subreddit]: [post]\\

Format your response starting with Classification: Can be ethically studied in a RCT or Classification: Cannot be ethically studied in a RCT

\normalfont

\section{Retrieval Configurations}
\label{appendix:retrieval}
We evaluated several retrieval configurations using the \verb|dunzhang/stella_en_400M_v5| embedding model on a small development set, with relevance judgments provided by a medical expert. Aggregate results are shown in Table~\ref{table:retrieval_strategies}. At the time of experimentation (09/11/2024), \verb|stella_en_400M_v5| was among the top-performing open-source embedding models on the HuggingFace MTEB benchmark while remaining computationally feasible for indexing approximately 800{,}000 RCT abstracts.

\begin{table*}[h]
\small 
    \centering
    \begin{tabular}{lc lc lc c c c c c}
    \toprule
    \textbf{Strategy} & \textbf{Query} & \textbf{Document} & \textbf{Pop.} & \textbf{Inter.} &\textbf{Out.} &\textbf{Overall} \\  \midrule
        S2P & Question PIO & PIO & 2.2 & 1.6 & 2 & 1.5 \\
        S2P & PIO & Abstract & 2.3 & 1.7 & 2.1 & 1.7 \\
        S2P & PIO & PIO & 2.3 & 1.7 & 2.1 & 1.6 \\
        \rowcolor{green}\textbf{S2S} & \textbf{PIO} & \textbf{Abstract} & \textbf{2.4} & \textbf{1.7} & \textbf{2.2} & \textbf{1.7}\\
        S2S & PIO & PIO \& Abstract & 2.2 & 1.6 & 2.0 & 1.6 \\
        S2S & PIO & PIO & 2.3 & 1.6 & 1.9 & 1.5 \\
        S2S & PIO & PIO \& Title \& Abstract & 2.2 & 1.6 & 2.0 & 1.6\\
        S2S & PIO & Title \& Abstract & 2.3 & 1.7 & 2.1 & 1.6\\
        \bottomrule
    \end{tabular}
    \caption{Retrieval strategy results on a test set of five claims. Scores report average relevance per claim, computed over 10 retrieved abstracts (1 = irrelevant, 2 = somewhat relevant, 3 = relevant). The highlighted row indicates the configuration used in the main study.}
    \label{table:retrieval_strategies}
\end{table*}

\section{Implementation Considerations for the Communication Model}
\label{appendix:implementation}
We emphasize that the communication model is an abstraction, and that the actual system could be implemented in many ways. The model could be a Reddit or Chrome extension tool for users to interact with while they write their Reddit post. Alternatively, the model could also publicly post follow-up questions as comments as a public reply to the users' post. The benefit of this approach is that there could be an expert in the loop that verifies the models' response, as well as these public responses could be helpful to other users reading the thread. We encourage the research community to explore various implementations of this system, as well as focus on extensive human and expert evaluation and systematic HITL methods.

\onecolumn
\newpage
\section{Pilot Claims}
\scriptsize

\begin{longtable}{p{0.97\textwidth}}
    \toprule
    \textbf{Post} \\
    \midrule
    \endfirsthead
    \toprule
    \textbf{Post} \\
    \midrule
    \endhead
    \midrule
    \multicolumn{1}{r}{\textit{Continued on next page}} \\
    \midrule
    \endfoot
    \bottomrule\\
        \caption{Pilot claims.}

    \endlastfoot
    \ClaimExample{microclots}{Could \hlpink{microclots} help explain the mystery of long Covid? \hlyellow{Acute }\hlblue{Covid-19}\hlyellow{ is not only a lung disease, but actually significantly affects the vascular (}\hlgreen{blood flow}\hlyellow{) and coagulation (}\hlgreen{blood clotting}\hlyellow{) systems. A connection to the damage done by }\hlblue{diabetes}\hlyellow{ might be possible.}}{r/Diabetes} \\
        \vspace{0.3em} \\
        \rowcolor{gray!10}

            \ClaimExample
        {Diabetes} 
        {Affording Medication

so im on a family plan with a 3k/6k out of pocket expense, I think it's hdhp with a hsa that my husband employer contributes a bit too. I know when I had coverage with my job i had a ppo plan. he's the one that chooses the plans at his job so im not the best when it comes to explaining the details for it..

was orginally taking metformin but it's horrible and over the past 2 months it's been making me sick as a dog so I asked my endocrinologist can I go on something else. she recommended \hlpink{ozempic}\hlyellow{ since alot of patients responded well to it, lost }\hlgreen{weight}\hlyellow{, and had a good effect on their }\hlgreen{sugar}\hlyellow{.} plus it's taken only weekly in which sounds great for someone like me since I'm not the best with keeping up with medications. back in December since we had hit our 6k deductible I had paid nothing when I recieced the medication so I had no clue what the actual price would be but I nearly had a heart attack when I tried picking it up in the store recently...with my plan I'm at 800 bucks for the thing and optum informed me it's 2300 (1981 with the discount card) for a 90 day supply. that's ALOT of money...I was going to purchase farxiga today with optum (1500 dollars) but literally don't have the money to afford to do so..my car needs a new catalytic converter so finacially I had to make the cut to my medication (my cardiologist put me on that to prevent heart failure since I have "resistant hypertension" that's not responding well medication)

i made a joke to my husband and said I may have to divorce him just so I qualify for that government health insurance. hell looking at it now I may be serious! as a diabetic or anyone in America on any type of medication how are ppl able to afford their insulin/pills/machines/ whatever. our household income is around 85k so there's not much assistance we can get that im aware of}{r/Diabetes} \\
\vspace{0.3em} \\
        \ClaimExample
        {Hallucinations} 
        {Do people with IH experience hallucinations?

I am so confused! My MSLT showed IH but my doctor gave me a clinical diagnosis of \hlblue{narcolepsy }because I experience hypnopompic \hlgreen{hallucinations} and \hlgreen{sleep paralysis}. \hlyellow{She told me people with IH dont experience those things which is why she switched the diagnosis.} Im confused because Ive read articles that say they are symptoms of IH. I know it doesnt really matter because treatment is the same, but I have this thing in me where I just need to know.}   
        {r/narcolepsy} \\
        \vspace{0.3em} \\
        \rowcolor{gray!10}
    \ClaimExample{COVID}{\hlblue{Epilepsy}\hlyellow{ Patients Much More Likely to Die of }\hlblue{COVID}
}{r/Epilepsy} \\
        \vspace{0.3em}\\
\ClaimExample{Long Covid}{\hl{I Had Never Felt Worse:}\hlpink{ Long Covid}\hlyellow{ Sufferers Are }\hlgreen{Struggling With Exercise}\hlyellow{ And experts have some theories as to why.} - The New York Times}{r/CFS} \\
        \vspace{0.3em} \\
        \rowcolor{gray!10}

\ClaimExample{Glycemic}{\hlpink{Dietary carbohydrate}\hlyellow{ restriction augments weight loss-induced improvements in glycaemic control and liver fat in individuals with type 2 }\hlblue{diabetes}: a randomised controlled trial. (Pub Date: 2022-01-07)
}{r/Diabetes} \\
        \vspace{0.3em} \\
\ClaimExample{Pfizer vaccine}{Pfizer vaccine

My son, non cf, is having his second pfizer \hlpink{vaccine}. He is 25 yrs old. For some reason I'm really nervous about it as \hlyellow{he has been told not to exercise for 48hrs afterwards due to heart inflammable young people are getting}...obvs this is rare...but my son is extremely active \& I'm in a tizz. He's having now as i write this. I'm extremely proud he is having it as alot of youngsters are refusing it atm but the anxiety over it is making me feel sick.}{r/CysticFibrosis} \\
\vspace{0.3em} \\
        \rowcolor{gray!10}

\ClaimExample{mold}{Mold and RA

I'm having a bit of a weird issue with \hlgreen{mold}. I'm currently in the process of being diagnosed with \hlblue{RA}. I've got \hlgreen{achy joints}, \hlgreen{swelling} whole nine yards. I transferred job locations earlier this month and was starting to feel better and my hand swelling finally went down. I then signed up for some overtime in my old job location and after about 2 hours my elbows and hands started to \hlgreen{ache} and swell. Every time it rains at this building water runs through the walls. I'm certain theirs mold in the walls. \hlyellow{Google says long-term toxic mold exposure can mimic RA}. Had anyone else had an experience with RA symptoms not ending up being RA or having one large trigger to RA symptoms. After going home and sleeping on things my hands started to feel better but not completely. }{r/rheumatoidarthritis} \\
\vspace{0.3em} \\

\ClaimExample{Copaxone}{copaxone vs aubagio?

My gf is about to switch from once a day copaxon injections to aubagio at the advice of her new neurologist.

After doing some research before starting the treatment, she is a bit worried about the liver function concerns with the drug.

My gf is bipolar, has high anxiety, and is on several meds for her mental health. \hl{I just pulled up a site that compared these 2 drugs and was really angry to see that }\hlpink{copaxone}\hlyellow{ patients reported it caused }\hlgreen{depression}\hlyellow{, }\hlgreen{anxiety}\hlyellow{, and other things the doctor never mentioned}. So I am cautiously optimistic that the change is in her best interest.

Any thoughts or experiences would be greatly appreciated. My research seems to lean towards the new medication, but we are obviously concerned at least about the liver function monitoring.

tyvm}{r/MultipleSclerosis} \\
\vspace{0.3em} \\
        \rowcolor{gray!10}

\ClaimExample{Calcium}{Calcium supplements (Citracal slow release) and total thyroidectomy
Hey all, I'm almost 6 years post total thyroidectomy, and since my providers at the time of my TT didn't really share any of this info/it was hard to track down, I wanted to put it out there for others.

First-- \hl{you'll probably want to get on a }\hlpink{calcium supplemen}\hlyellow{t}. That part I was told. How much calcium I wasn't told, but eventually found out from a pharmacist to go a bit above the recommended for your age/assigned sex at birth. Normally my recommended would be 1000mg, but because of the TT, it's 1200mg.

Second-- wait at least 4 hours after taking your thyroid hormone replacement before taking a calcium supplement. Also was told that, also something everyone here probably already knows.

Third-- our bodies can only absorb around 500mg of calcium in one go.

Fourth-- if you've had a TT, your body will absorb calcium citrate more effectively than calcium carbonate. I learned this literally a month ago from a PCP who doesn't specialize in thyroid health, and I'd love to know why my endocrinologist never told me.

Now, all of the above led me to be interested in the Citracal slow release, as it's 1200mg, but released slowly so you can take it once a day and still get all of it. My only issue was that I couldn't find anywhere that said how long it took to fully release. I was worried that if it took too long, it would prevent my levothyroxine from absorbing the next day. I couldn't find the answer online, but finally called their questions line today and found out it's 8 hours.

Obviously I'm not a medical provider, I just want more of us to have access to this info, especially since a lot of us have worked with medical providers that don't give us all of the information we need. Also not here to advertise for that brand specifically, I just wanted something convenient enough to take once a day, and figured others might have had the same question!}{r/thyroidcancer} \\\vspace{0.3em} 
\label{table:pilotclaims}
\end{longtable}

\label{appendix:pilot_section}

\section{Refinement Claims}
\label{appendix:refinement_section}
\scriptsize

\begin{longtable}{p{0.97\textwidth}}
    \toprule
    \textbf{Post} \\
    \midrule
    \endfirsthead
    \toprule
    \textbf{Post} \\
    \midrule
    \endhead
    \midrule
    \multicolumn{1}{r}{\textit{Continued on next page}} \\
    \midrule
    \endfoot
    \bottomrule \\
        \caption{}

    \endlastfoot
            \ClaimExample
        {Ivabradine}
        {Anyone here with low \hlblue{bp} take \hlpink{Ivabradine}?
        
Im just wanting to do a bit of research on different meds before my doctors appointment. Last time they told me they cant medicate me because my bp dropped pretty low during the TTT. However, at rest my bp is normal and even standing up it doesnt drop noticeably low unless Im standing still for a longer period of time.

So I just wanted to know if any of yall are in a similar situation and have good (or bad) experiences with this drug. \hlyellow{I hear }\hlpink{midodrine}\hlyellow{ is good for low bp but its more expensive and the side effects sound kind of iffy to me.}
}{r/POTS} \\
        \vspace{0.3em} \\
        \rowcolor{gray!10}
            \ClaimExample
        {Fluoxetine} 
        {Long-lasting apathetic tendencies, anhedonia etc.

I'm just apathetic in general, and am unable to do even the smallest things. \hlpink{Fluoxetine}\hlyellow{ might have had some positive effects}, and I'm supposed to be taking it now, but I can't even be bothered to get a refill.

I can't tell whether or not my asocial tendencies are a personality trait. I currently have no interest in maintaining a relationship with family or friends.

I've never been diagnosed with dysthymia - only depression - but a lot of the symptoms seem relevant, and my doctor did mention it at one point.}   
        {r/Dysthymia} \\
                \vspace{0.3em} \\

    \ClaimExample{Psychosis and Antidepressants}{\hlblue{Psychosis} and \hlpink{antidepressants}

Hey everyone!

So some crazy stuff happened to me over the last week. I am on abilify for my psychosis and I have been suffering from \hlgreen{depression}.

My doctor decided to prescribe me \hlpink{Wellbutrin} 150mg first. Took it for about five days, started having extreme anxiety and \hlgreen{dry mouth}. I mentioned this to my doctor and he switched me to \hlpink{Lexapro} 5mg. Extreme anxiety and dry mouth but something new happened this time -my fucking \hlgreen{delusions} and \hlgreen{hallucinations} came back. I had to legit tell myself my thoughts werent based on reality. But holy crap was it difficult. I didnt take any antidepressants today and already feel better.

This is crazy, has anyone experienced anything like this? \hlyellow{I didnt think anti depressants would bring out my psychosis.} Guess I might have to go the natural route for my depression :(
}{r/Psychosis} \\        \vspace{0.3em} \\
        \rowcolor{gray!10}

\ClaimExample{Prednisone}{Cytoxan and prednisone

Rheumatologist says \hlpink{cellcept} failed to protect my kidneys and now I have developed \hlgreen{lupus nephritis}.Im so upset. \hlpink{Prednisone} messed up my hips so badly that they both need to be replacedI dont want to get back on it but \hlyellow{rheumatologist says its to bring the }\hlgreen{inflammation}\hlyellow{ down in my kidneys.} Ive never been on \hlpink{Cytoxan} but the side effects sound identical to a lupus flare. How am I supposed to be positive with news like this? I feel so defeatedI dont know what to do.}{r/lupus}  \\       \vspace{0.3em} \\

\ClaimExample{Metformin replace Insulin}{Can metformin replace insulin?

I realize this is definitely case-by-case but Im curious to know if anyone has been able to get off of insulin and take just metformin? When I was diagnosed with type 2 I was automatically put on insulin and generally take quite a bit of it, but \hl{now after some research Im considering asking my doctor to try treatment through }\hlpink{metformin }\hlyellow{in February}.
}{r/Diabetes} \\ \vspace{0.3em} 

\end{longtable}

\normalsize
\twocolumn
\section{License Information}
\label{app:license}

We predominantly used the RedHOT dataset and abstracts from the TrialStreamer database in our work.
Both of these works are licensed on a Creative Commons Attribution 4.0 International License.

We will also release our work under a Creative Commons Attribution 4.0 International License.

\end{document}